\RequirePackage{fix-cm}
\documentclass[twocolumn]{svjour3}          
\smartqed  
\usepackage{graphicx}

\usepackage{amssymb}
\usepackage{tikz}
\usetikzlibrary{shapes, arrows.meta}
\usetikzlibrary{backgrounds}
\usepackage{pagecolor,lipsum}
\usepackage{multirow}
\usepackage{xspace}
\usepackage{booktabs}
\usepackage{doi}

\def\makeheadbox{\relax}
\makeatletter
\let\@date\relax
\makeatother
\usepackage{dirtytalk}
\usetikzlibrary{arrows, decorations.markings}
\tikzset{snake it/.style={-stealth,
decoration={snake, 
    amplitude = .4mm,
    segment length = 2mm,
    post length=0.9mm},decorate}}

\tikzstyle{vecArrow} = [thick, decoration={markings,mark=at position
   1 with {\arrow[semithick]{open triangle 60}}},
   double distance=1.4pt, shorten >= 5.5pt,
   preaction = {decorate},
   postaction = {draw,line width=1.4pt, white,shorten >= 4.5pt}]
\tikzstyle{innerWhite} = [semithick, white,line width=1.4pt, shorten >= 4.5pt]

\tikzstyle{vecArrow*} = [thick, decoration={markings,mark=at position
   1 with {\arrow[semithick]{open triangle 60}}},
   double distance=1.4pt, shorten >= 5.5pt,
   preaction = {decorate},
   postaction = {draw,line width=1.4pt, white,shorten >= 4.5pt}, 
   to path={-- node[inner sep=5pt,at end,sloped] {\huge${}^*$} (\tikztotarget) \tikztonodes}]
\definecolor{nodeblue}{rgb}{0.5,0.7,1}
\definecolor{nodered}{rgb}{1,0.4,0.4}
\definecolor{nodepink}{rgb}{0.97,0.43,0.97}
\definecolor{edgeblue}{rgb}{0.2,0.4,1}
\definecolor{edgered}{rgb}{1,0.2,0.2}
\makeatletter
\newcommand{\miniscule}{\@setfontsize\miniscule{7}{8}}
\makeatother

\usepackage{array}
\newcolumntype{L}[1]{>{\raggedright\let\newline\\\arraybackslash\hspace{0pt}}m{#1}}
\newcolumntype{C}[1]{>{\centering\let\newline\\\arraybackslash\hspace{0pt}}m{#1}}
\newcolumntype{R}[1]{>{\raggedleft\let\newline\\\arraybackslash\hspace{0pt}}m{#1}}

\newcommand{\ttt}{\texttt}

\usepackage[colorinlistoftodos]{todonotes}

\begin{document}

\title{Evolving Graphs with Semantic Neutral Drift
}


\author{Timothy Atkinson         \and
        Detlef Plump \and
        Susan Stepney
}


\institute{Department of Computer Science, University of York, UK\\
              \email{$\{$tja511,detlef.plump,susan.stepney$\}$@york.ac.uk}  
}

\date{}

\maketitle

\begin{abstract}
We introduce the concept of Semantic Neutral Drift (SND) for genetic programming (GP), where we exploit equivalence laws to design semantics preserving mutations guaranteed to preserve individuals' fitness scores. A number of digital circuit benchmark problems have been implemented with rule-based graph programs and empirically evaluated, demonstrating quantitative improvements in evolutionary performance. Analysis reveals that the benefits of the designed SND reside in more complex processes than simple growth of individuals, and that there are circumstances where it is beneficial to choose otherwise detrimental parameters for a GP system if that facilitates the inclusion of SND.
\end{abstract}

\section{Introduction}
\label{intro}
In genetic programming the ability to escape local optima is key to finding globally optimal solutions. 
Neutral drift, a mechanism whereby individuals with fitness-equivalent phenotypes to the existing population may be generated by mutation \cite{galvan2011neutrality}
offers the search of new neighborhoods for sampling thus increasing the chance of leaving local optima. A number of studies on neutrality in Cartesian Genetic Programming (CGP) \cite{miller2006redundancy,vassilev2000advantages,Turner2015} find it to be an almost always beneficial property for studied problems. In general, comparative studies \cite{Miller2011-CGP} find that CGP using only mutation and neutral drift is able to compete with traditional tree-based Genetic Programming (GP) which uses more familiar crossover operators (see \cite{koza:book}) to introduce genetic variation. 

\cite{Turner2015} makes a distinction between \textit{implicit} neutral drift (where a genetic operator yields a semantically equivalent child) and \textit{explicit} neutral drift (where a genetic operator only modifies intronic code). We note that many comparative studies largely focus on the role of both types of neutral drift as byproducts of existing genetic operators and neutrality within the representation \cite{miller2006redundancy,vassilev2000advantages,Turner2015,banzhaf1994genotype} rather than as deliberately designed features of an evolutionary system. We propose the opposite; to employ domain knowledge of equivalence laws to specify mutation operators on the active components of individuals which always induce neutral drift. Hence our work can be viewed as an attempt to explicitly induce additional implicit neutral drift in the sense of \cite{Turner2015}.

We build on our approach EGGP (Evolving Graphs by Graph Programming) \cite{Atkinson-Plump-Stepney18a}, by implementing \textit{semantics preserving mutations} to directly achieve neutral drift on the active components of individual solutions. Here, we implement logical equivalence laws as mutations on the active components of candidate solutions to digital circuit problems to produce semantically equivalent,  equally fit, children. While our semantics-preserving mutations produce semantically equivalent children they do not guarantee preservation of size; our fitness measures evaluate semantics only, not, for example, size or complexity.

We describe and implement Semantic Neutral Drift straightforwardly by using rule-based graph programs, here in the probabilistic language P-GP\,2 \cite{Atkinson-Plump-Stepney18b}. This continues from \cite{Atkinson-Plump-Stepney18a} where we use a probabilistic variant of the graph programming language GP\,2 to design acylicity-preserving edge mutations for digital circuits that correctly identify the set of all possible valid mutations. The use of P-GP\,2 here enables concise description of complex transformations such as DeMorgan's laws by identifying and rewriting potential matches for these laws in the existing formalism of graph transformation. This reinforces the notion that the direct encoding of solutions as graphs is useful as it allows immediate access to the phenotype of individual solutions and makes it possible to design complex mutations by using powerful algorithmic concepts from graph programming.

We investigate four sets of semantics-preserving mutations for digital circuit design, three built upon logical equivalence laws and a fourth taken from term-graph rewriting. We run EGGP with each rule-set on a set of benchmark problems and establish statistically significant improvements in performance for most of our visited problems. An analysis of our results reveals evidence that it is the semantic transformations, beyond simple `neutral growth', which are aiding performance. We then combine our two best performing sets of mutation operators and evaluate this new set under the same conditions, achieving further improvements. We also provide evidence that, although operators implementing semantics-preserving mutations may be more difficult to use, the inclusion of those semantics-preserving mutations may allow evolution to out-perform equivalent processes that use `easier' operators.

The rest of this paper is organised as follows. In Section \ref{sec:neutral} we review existing literature on Neutral Drift in Genetic Programming. In Sections \ref{sec:GP2} and \ref{sec:EGGP} we describe the graph programming language GP\,2 and our existing approach EGGP. In Section \ref{sec:NeutralEGGP} we describe our extension to EGGP where we incorporate deliberate neutral drifts into the evolutionary process. In Section \ref{sec:Exp} we describe our experimental setup and in Section \ref{sec:Results} we give the results from these experiments. In Section \ref{sec:Analysis} we provide in-depth analysis of these results to establish precisely what components of our approach are aiding performance. In Section \ref{sec:Conclusion} we conclude our work and propose potential future work on this topic.

\section{Neutral Drift in Genetic Programming \label{sec:neutral}}

Neutral drift remains a controversial subject in Evolutionary Computation. See \cite{galvan2011neutrality} for a survey. Here, we focus on neutrality in the context of genetic programming as the most relevant area to our own work; there is also literature on, for example, genetic algorithms \cite{HarveyT96} and landscape analysis \cite{barnett1998ruggedness}. 

The process of neutral drift might be described as the mutation of individual candidate solutions to a given problem without advantageous or deleterious effect on their fitness. This exposes the evolutionary algorithm to a fitness `plateau' with each fitness-equivalent individual offering a different portion of the landscape to sample. Neutral drift can be viewed as random walks on the neighborhoods of surviving candidate solutions. 
In a system with neutral drift, an apparently local optimum might be escaped by `drifting' to some other fitness-equivalent solution that  has advantageous mutations available.

The most apparent demonstration of neutral drift in genetic programming literature occurs in Cartesian Genetic Programming (CGP) \cite{Miller2000-CGP}. In CGP, individuals encode directed acyclic graphs; some portion of a genome may be `inactive', contributing nothing to the phenotypic fitness, because it represents a subgraph that is not connected to the phenotype's main graph. These inactive genes can mutate without influencing an individual's fitness and then, at some later point, may become active. Early work on CGP has found that by allowing neutral drift to take place (by choosing a fitness-equivalent child over its parent in the $1+\lambda$ algorithm), the success rate of experiments significantly improves \cite{vassilev2000advantages}. A later claim that neutrality in CGP aids search in needle-in-haystack problems \cite{yu2002} has been contested by a counter-claim that better performance can be achieved by random search \cite{collins2006finding}. \cite{miller2006redundancy} finds that better performance can be achieved with neutral drift enabled by increasing the amount of redundant material present in individuals. \cite{Turner2015} establishes a distinction between \textit{explicit} and \textit{implicit} neutral drift. Explicit neutral drift occurs on inactive components of the individual, whereas implicit neutral drift occurs when active components of the individual are mutated but the fitness does not change. The authors were able to isolate explicit neutral drift and demonstrate that it offers additive benefits beyond those of implicit neutral drift.

Outside of CGP, \cite{banzhaf1994genotype} proposes a form of Linear Genetic Programming where programs are decoded from bit-strings, and redundancy exists, in that certain operations have multiple representations. A study of evolvability in Linear GP \cite{Hu2009} found that neutrality cooperates with `variability' (the ability of a system to generate phenotypic changes) to generate adaptive phenotypic changes which aid the overall ability of the system to respond to the landscape. Recent work \cite{Hu2018} studying the role of neutrality in small Linear GP programs found that the robustness of a genotype (the proportion of its neighbours within the landscape which are neutral changes) has a complex and non-monotonic relationship with the overall evolvability of the genotype. 

In \cite{downing2005evolving}, binary decision diagrams are evolved with explicit neutral mutations. Although those neutral mutations are not isolated for their advantages/disadvantages, a later work has found that a higher rate of neutral drift on binary decision diagrams is advantageous \cite{downing2006neutrality}. Koza also makes some reference to the ideas we employ in Section \ref{sec:NeutralEGGP} when he describes the editing digital circuits by applying DeMorgan's laws to them  \cite[Ch.6]{koza:book}. A study of neutrality in tree-based GP for boolean functions \cite{VANNESCHI201234} found a correlation between using a more effective function set and the existence of additional neutrality when using that function set.

While not directly related to neutrality, a number of investigations have been carried out exploring the notion of semantically aware genetic operators to improve the locality of mechanisms such as crossover in tree-based GP \cite{Moraglio2012,10.1007/978-3-642-01181-8_25}. We refer the reader to the extensive survey \cite{Vanneschi2014} on this field of research. Whereas neutrality is the process whereby phenotypically identical and genotypically distinct individuals are visited by the evolutionary process, semantically aware genetic operators attempt to produce phenotypically 'close' individuals to improve the locality of the search neighbourhood. It should be noted that employing semantically aware genetic operators may sometimes lead to a loss of diversity \cite{Pham2013}. It could be argued that the deliberate neutral operators we propose in this work are a form of semantically aware mutation operators designed to explicitly exploit neutrality.

Neutral drift has some parallels with work on biological evolution. Kimura's \textit{Neutral Theory of Molecular Evolution} \cite{kimura1983neutral} posits that most mutations in nature are neither advantageous or deleterious, instead introducing `neutral' changes that do not affect phenotypes but account for much of the genetic variation within and between species. While Kimura's theory remains controversial (see \cite{hahn2008toward}), it appears to loosely correspond to the notions of neutral mutation described in genetic programming literature.

Throughout the literature we have covered, neutrality is mostly considered in the sense of \textit{explicit neutral drift} as defined in \cite{Turner2015}. Conversely in our work here we are focusing on neutral drift on the active components of individual solutions, with some relationship therefore to the neutral mutations on binary decision diagrams in \cite{downing2005evolving}.

\section{Graph Programming with P-GP\,2 \label{sec:GP2}}

Here we give a brief introduction to the graph programming language GP\,2; see \cite{Plump17a} for a detailed account of the syntax and semantics of the language. 

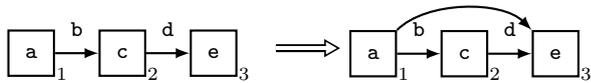
\begin{figure}[tb]
\texttt{Main} := \texttt{link}!\\
\,\\
$\texttt{link(a,b,c,d,e:list)}$

\centering
\begin{tikzpicture}
\begin{scope}[node distance=4.5cm, every node/.style={rectangle,thick,draw=white,text=black,minimum size = 0.6cm}]
    \node (K) {
      \begin{tikzpicture}
      \begin{scope}[node distance=1.2cm, every node/.style={rectangle,thick,draw}]
          \node[label={[xshift=0.4cm, yshift=-0.9cm]:\miniscule1}] (A) at (0, 0) {$\texttt{a}$};
          \node[label={[xshift=0.4cm, yshift=-0.9cm]:\miniscule2},right of = A] (B) {$\texttt{c}$};
          \node[label={[xshift=0.4cm, yshift=-0.9cm]:\miniscule3}, right of = B] (C)  {$\texttt{e}$};
      \end{scope}

      \begin{scope}[>={Stealth[black]},
                    every node/.style={fill=white,rectangle,inner sep=0,fill=none},
                    every edge/.style={thick}]
                      \path [->, -latex] (A) edge[draw] node[above=0.3pt,inner sep=1pt] {$\texttt{b}$} (B);
                      \path [->, -latex] (B) edge[draw] node[above=0.3pt,inner sep=1pt] {$\texttt{d}$} (C);
                      \path [->, -latex] (A) edge[above=0.3pt,in=135,out=45,looseness=0.8,draw=none] (C);
      \end{scope}
      \end{tikzpicture}
 };
 
    \node[right of=K] (R){
      \begin{tikzpicture}
      \begin{scope}[node distance=1.2cm, every node/.style={rectangle,thick,draw}]
          \node[label={[xshift=0.4cm, yshift=-0.9cm]:\miniscule1}] (A) at (0, 0) {$\texttt{a}$};
          \node[label={[xshift=0.4cm, yshift=-0.9cm]:\miniscule2},right of = A] (B) {$\texttt{c}$};
          \node[label={[xshift=0.4cm, yshift=-0.9cm]:\miniscule3}, right of = B] (C)  {$\texttt{e}$};
      \end{scope}

      \begin{scope}[>={Stealth[black]},
                    every node/.style={fill=white,rectangle,inner sep=0,fill=none},
                    every edge/.style={draw,thick}]
                      \path [->, -latex] (A) edge node[above=0.3pt,inner sep=1pt] {$\texttt{b}$} (B);
                      \path [->, -latex] (B) edge node[above=0.3pt,inner sep=1pt] {$\texttt{d}$} (C);
                      \path [->, -latex] (A) edge[above=0.3pt,in=135,out=45,looseness=0.8] (C);
      \end{scope}
      \end{tikzpicture}
 };
\end{scope}

  \draw[vecArrow] (K) to (R);

  \draw[innerWhite] (K) to (R);
\end{tikzpicture}

\leftskip=0pt$\texttt{where not edge(1,3)}$
\caption{\label{fig:transitive-closure} A GP\,2 program computing the transitive closure of a graph.} 
\end{figure}

A graph program consists of declarations of \emph{graph transformation rules} and a main command sequence controlling the application of the rules. Graphs are directed and may contain loops and parallel edges. The rules operate on \emph{host graphs}\/ whose nodes and edges are labelled with integers, character strings or lists of integers and strings. Variables in rules (relevant for this paper) are of type \texttt{int}, \texttt{string} or \texttt{list}. Integers and strings are considered as lists of length one, hence every label in GP\,2 is a list. For example, in Figure \ref{fig:transitive-closure}, the list variables \texttt{a}, \texttt{c} and \texttt{e} are used as node labels while \texttt{b} and \texttt{d} serve as edge labels. The small numbers attached to nodes are identifiers that specify the correspondence between the nodes in the left and the right graph of the rule.

Besides carrying list expressions, nodes and edges can be \emph{marked}. For example, in the program of Figure \ref{fig:dac_mutate}, blue and red node marks are used to prevent the rule \texttt{mutate$\_$edge} from creating a cycle. In rules, a magenta colour can be used as a wildcard for any mark. For example, in the rules \texttt{remove\_edge}, \texttt{unmark\_edge} and \texttt{unmark\_node} of Figure \ref{fig:seman}, pairs of magenta nodes with the same identifier on the left and the right represent nodes with the same green, blue or grey mark.

The principal programming constructs in GP\,2 are conditional graph-trans\-for\-mation rules labelled with expressions. To apply a rule to a host graph, the rule is first instantiated by replacing all variables with values and evaluating the expressions. The rule's condition, if present, has to evaluate to true. Then the left graph of the instantiated rule is matched (injectively) with a subgraph of the host graph. Finally the subgraph is replaced with the right graph of the instantiated rule. This means that the nodes corresponding to the numbered nodes of the left graph are preserved (but possibly re-labelled), any other nodes and all edges of the left graph are deleted, and any unnumbered nodes and all edges of the rule's instantiated right graph are inserted.

For example, given any host graph $G$, the program in Figure \ref{fig:transitive-closure} produces the smallest transitive graph that results from adding unlabelled edges to $G$. (A graph is \emph{transitive} if for each directed path from a node $v_1$ to another node $v_2$, there is an edge from $v_1$ to $v_2$.) 
The program applies the single rule \texttt{link} \emph{as long as possible} to a host graph. In general, any subprogram can be iterated with the postfix operator ``\texttt{!}''. Applying \texttt{link} amounts to non-deterministically selecting a subgraph of the host graph that matches \texttt{link}'s left graph, and adding to it an edge from node 1 to node 3 provided there is no such edge (with any label). The application condition \texttt{where not edge(1,3)} ensures that the program terminates and extends the host graph with a minimal number of edges. 

Besides applying individual rules, a program may apply a rule set $\mathtt{\{}r_1,\dots,r_n\mathtt{\}}$ to the host graph by non-deterministically selecting a rule $r_i$ among the applicable rules and applying it. Further control constructs include the sequential composition $P\mathtt{;}\, Q$ of programs $P$ and $Q$, and the branching construct \ttt{try} $T$ \ttt{then} $P$ \ttt{else} $Q$. To execute the latter, test $T$ is executed on the host graph $G$ and if this results in some graph $H$, program $P$ is executed on $H$. If $T$ fails (because a rule or set of rules cannot be matched), program $Q$ is executed on $G$. The variant \ttt{try} $T$ of this construct executes $T$ on $G$ and if this results in graph $H$, returns $H$. If the execution fails, $G$ is returned unmodified.

In general, the execution of a program on a host graph may result in different graphs, fail, or diverge.  The \emph{semantics}\/ of a program $P$ maps each host graph to the set of all possible outcomes \cite{Plump12a}. GP\,2 is computationally complete in that every computable function on graphs can be programmed \cite{Plump17a}. 

GP\,2's inherent non-determinism is useful as many graph problems are naturally multi-valued, for example the computation of a shortest path or a minimum spanning tree. The results described in the rest of this paper have been obtained with a probabilistic extension of GP\,2, called P-GP\,2. This provides a rule-set command $\mathtt{[r_1,\dots,r_n\mathtt]}$ which chooses a rule uniformly at random among the applicable rules and applies the rule with a match selected uniformly at random among all matches of that rule \cite{Atkinson-Plump-Stepney18b}.

\section{Evolving Graphs by Graph Programming (EGGP) \label{sec:EGGP}}

\subsection{Introduction to EGGP}

In \cite{Atkinson-Plump-Stepney18a} we introduce EGGP, an evolutionary algorithm that evolves graphs (specifically, in that case, digital circuits) using graph programming. We have found that by evolving graphs directly and designing mutation operators that respect the constraints of the problem, we are able to significantly outperform CGP under similar conditions on a number of digital circuit benchmark problems. In this section we formally describe this approach.

Our approach is justified by two observations: 
(i)~the use of graphs as a representation is beneficial, as it directly addresses a number of motivating problems within computer science such as neural network topology, Bayesian network topology, digital circuit design, program design, and quantum circuit design; (ii) with graphs as a representation it is necessary to have a language to describe the neighborhoods (mutations) on individuals. Graph programming readily lends itself to this endeavour due to its computational completeness over functions on graphs.

Our approach is not alone in addressing the issue of evolving graphs and graph-like programs. CGP \cite{Miller2000-CGP}, where individuals encode directed acyclic graphs, is a primary candidate for related work and is used as a benchmark here. Parallel Distributed Genetic Programming \cite{poli1997evolution,Poli1999} introduces a `graph on a grid' representation for genetic programming in a similar manner to the grid-like description of CGP, allowing the evolution of programs with multiple outputs and sharing. MIOST \cite{LopezR07} also extends traditional genetic programming to these same concepts of multiple outputs and sharing. For a more detailed discussion of related approaches, see \cite{Atkinson-Plump-Stepney18a}. Our approach differs from these in that (i) we deal with graphs directly rather than through an encoding or some subset of graphs; and (ii) our mutation operators are domain-specific and may be changed to suit the constraints of a problem and to exploit domain-specific knowledge.

Here we address the problems of digital circuits, primarily because they suit our discussion of neutral drift by design. For this reason, the rest of this paper focuses on the evolution of digital circuits as a concrete case study.

\subsection{Evolving Digital Circuits as Graphs}

\begin{figure}[!t]
\centering
\begin{tikzpicture}
\begin{scope}[every node/.style={rectangle,thick,draw,minimum size=0.6cm,}]
    \node[align=center] (1) at (-1, -2) {\miniscule$\texttt{"IN"}:0$};
    \node[align=center] (2) at (1, -2) {\miniscule$\texttt{"IN"}:1$};
    \node[align=center] (3) at (-2, -0.5) {\miniscule$\texttt{"AND"}:2$};
    \node[align=center] (4) at (0, -0.5) {\miniscule$\texttt{"OR"}:2$};
    \node[align=center] (5) at (2,-0.5) {\miniscule$\texttt{"OR"}:2$};
    \node[align=center] (6) at (-3, 1) {\miniscule$\texttt{"NOT"}:1$};
    \node[align=center] (7) at (-1, 1) {\miniscule$\texttt{"NOT"}:1$};
    \node[align=center] (8) at (1, 1) {\miniscule$\texttt{"AND"}:2$};
    \node[align=center] (9) at (3, 1) {\miniscule$\texttt{"AND"}:2$};
    \node[align=center] (10) at (-2, 2.5) {\miniscule$\texttt{"AND"}:2$};
    \node[align=center] (11) at (0, 2.5) {\miniscule$\texttt{"OR"}:2$};
    \node[align=center] (12) at (2, 2.5) {\miniscule$\texttt{"NOT"}:1$};
    \node[align=center] (13) at (-1, 4) {\miniscule$\texttt{"OUT"}:0$};
    \node[align=center] (14) at (1, 4) {\miniscule$\texttt{"OUT"}:1$};
\end{scope}

\begin{scope}[>={Stealth[black]} ,
                    every node/.style={fill=white,circle},
                    every edge/.style={draw,thick}]
              \path [->, -latex] (13) edge  (7);
              \path [->, -latex] (14) edge  (12);
              \path [->, -latex] (12) edge  (8);
              \path [->, -latex] (11) edge  (8);
              \path [->, -latex] (11) edge  (7);
              \path [->, -latex] (10) edge  (7);
              \path [->, -latex] (10) edge  (11);
              \path [->, -latex] (9) edge  (8);
              \path [->, -latex] (9) edge  (5);
              \path [->, -latex] (8) edge  (7);
              \path [->, -latex] (8) edge  (4);
              \path [->, -latex] (7) edge  (4);
              \path [->, -latex] (6) edge  (3);
              \path [->, -latex] (5) edge  (4);
              \path [->, -latex] (5) edge  (2);
              \path [->, -latex] (4) edge  (2);
              \path [->, -latex] (4) edge  (1);
              \path [->, -latex] (3) edge  (1);
              \path [->, -latex] (3) edge  (2);
\end{scope}
\end{tikzpicture}
\caption{A P-GP\,2 encoding of a 2-input, 2-output digital circuit over the function set $\{\texttt{AND}, \texttt{OR}, \texttt{NAND}, \texttt{NOR}\}$. 
(The \textit{outgoing} edges of a node point to the sources of that node's \textit{input values}, following  the convention used in the graph programming community).
Output 0 (corresponding to the node labelled $\texttt{"OUT"}:0$) has logical behaviour $\neg(i_0 \lor i_1)$ where $i_0$ and $i_1$ correspond to the input nodes labelled $\texttt{"IN"}:0$ and $\texttt{"IN"}:1$ respectively. \label{fig:dac_individual}}
\end{figure}
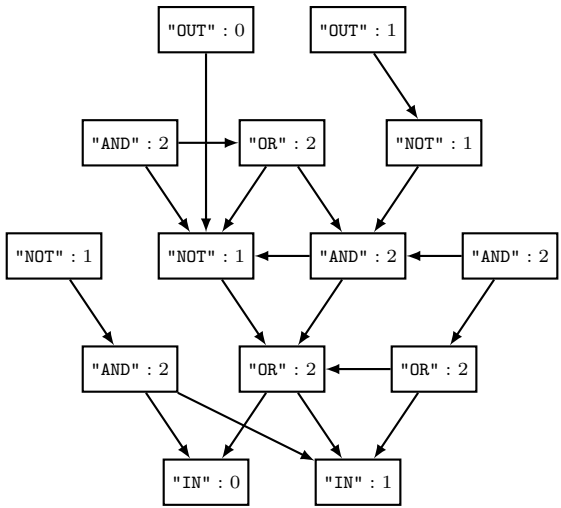

\begin{figure*}[!t]
\begin{center}
 \begin{tabular}{@{}l@{}}
\small\texttt{Main} := \texttt{try} (\texttt{[pick\char`_edge]}; \texttt{mark\char`_output}!; \texttt{[mutate\char`_edge]}; \texttt{unmark}!)\vspace{5.0mm}\\
\centering
\begin{tabular}{p{6cm}c|cp{6cm}}
\small$\texttt{pick\char`_edge(a,b,c:list)}$ 
\medbreak
\centering
\begin{tikzpicture}
\begin{scope}[node distance=3.5cm, every node/.style={rectangle,minimum size=0.6cm,thick,draw=white,text=black}]
    \node (K) {
      \begin{tikzpicture}
      \begin{scope}[node distance=1.1cm, every node/.style={thick,draw}]
          \node[label={[xshift=0.4cm, yshift=-0.9cm]:\miniscule1}] (A) {\small$\texttt{a}$};
          \node[label={[xshift=0.4cm, yshift=-0.9cm]:\miniscule2},right of =A] (B) {\small$\texttt{c}$};
      \end{scope}

      \begin{scope}[>={Stealth[black]},
                    every node/.style={fill=white,rectangle,inner sep=0,fill=none},
                    every edge/.style={draw,thick}]
                      \path [->, -latex] (A) edge node[above=0.5pt,inner sep=1pt]  {\small$\texttt{b}$} (B);
      \end{scope}
      \end{tikzpicture}
 };
 
    \node[right of=K] (R){
      \begin{tikzpicture}
      \begin{scope}[node distance=1.1cm, every node/.style={thick,draw}]
          \node[label={[xshift=0.4cm, yshift=-0.9cm]:\miniscule1},fill = nodeblue] (A) {\small$\texttt{a}$};
          \node[label={[xshift=0.4cm, yshift=-0.9cm]:\miniscule2},right of =A, fill = nodered] (B) {\small$\texttt{c}$};
      \end{scope}

      \begin{scope}[>={Stealth[black]},
                    every node/.style={fill=white,rectangle,inner sep=0,fill=none},
                    every edge/.style={draw,thick}]
                      \path [->, -latex, draw = edgered, edgered] (A) edge node[above=0.5pt,inner sep=1pt] {\small$\texttt{b}$} (B);
      \end{scope}
      \end{tikzpicture}
 };
\end{scope}
  \draw[vecArrow] (K) to (R);

  \draw[innerWhite] (K) to (R);
\end{tikzpicture}

\bigbreak
\leftskip=0pt\small$\texttt{mark\char`_output(a,b,c:list)}$
\medbreak
\centering
\begin{tikzpicture}
\begin{scope}[node distance=3.5cm, every node/.style={rectangle,minimum size=0.6cm,thick,draw=white,text=black}]
    \node (K) {
      \begin{tikzpicture}
      \begin{scope}[node distance=1.1cm, every node/.style={thick,draw}]
          \node[label={[xshift=0.4cm, yshift=-0.9cm]:\miniscule1}] (A) {\small$\texttt{a}$};
          \node[label={[xshift=0.4cm, yshift=-0.9cm]:\miniscule2},right of =A,fill = nodeblue] (B) {\small$\texttt{c}$};
      \end{scope}

      \begin{scope}[>={Stealth[black]},
                    every node/.style={fill=white,rectangle,inner sep=0,fill=none},
                    every edge/.style={draw,thick}]
                      \path [->, -latex] (A) edge node[above=0.5pt,inner sep=1pt] {\small$\texttt{b}$} (B);
      \end{scope}
      \end{tikzpicture}
 };
 
    \node[right of=K] (R){
      \begin{tikzpicture}
      \begin{scope}[node distance=1.1cm, every node/.style={thick,draw}]
          \node[label={[xshift=0.4cm, yshift=-0.9cm]:\miniscule1},fill = nodeblue] (A) {\small$\texttt{a}$};
          \node[label={[xshift=0.4cm, yshift=-0.9cm]:\miniscule2},right of =A, fill = nodeblue] (B) {\small$\texttt{c}$};
      \end{scope}

      \begin{scope}[>={Stealth[black]},
                    every node/.style={fill=white,rectangle,inner sep=0,fill=none},
                    every edge/.style={draw,thick}]
                      \path [->, -latex] (A) edge node[above=0.5pt,inner sep=1pt] {\small$\texttt{b}$} (B);
      \end{scope}
      \end{tikzpicture}
 };
\end{scope}
  \draw[vecArrow] (K) to (R);

  \draw[innerWhite] (K) to (R);
\end{tikzpicture} 

& \, & \,&

\small$\texttt{mutate\char`_edge(a,b,c,d:list; s:string)}$
\medbreak
\centering
\begin{tikzpicture}
\begin{scope}[node distance=3.5cm, every node/.style={rectangle,minimum size=0.6cm,thick,draw=white,text=black}]
    \node (K) {
      \begin{tikzpicture}
      \begin{scope}[node distance=1.1cm, every node/.style={thick,draw, text width=6mm, align=center,
  inner sep=0pt}]
          \node[label={[xshift=0.4cm, yshift=-0.9cm]:\miniscule1},fill = nodeblue] (A) {\small$\texttt{a}$};
          \node[label={[xshift=0.4cm, yshift=-0.9cm]:\miniscule2},below of =A, fill = nodered] (B) {\small$\texttt{b}$};
          \node[label={[xshift=0.4cm, yshift=-0.9cm]:\miniscule3},right of =A] (C) {\small$\texttt{s:c}$};
      \end{scope}

      \begin{scope}[>={Stealth[black]},
                    every node/.style={fill=white,rectangle,inner sep=0,fill=none},
                    every edge/.style={draw,thick}]
                      \path [->, -latex, draw = edgered, edgered] (A) edge node[xshift=2mm,inner sep=1pt]  {\small$\texttt{d}$} (B);
      \end{scope}
      \end{tikzpicture}
 };
 
    \node[right of=K] (R){
      \begin{tikzpicture}
      \begin{scope}[node distance=1.1cm, every node/.style={thick,draw, text width=6mm, align=center,
  inner sep=0pt}]
          \node[label={[xshift=0.4cm, yshift=-0.9cm]:\miniscule1}] (A) {\small$\texttt{a}$};
          \node[label={[xshift=0.4cm, yshift=-0.9cm]:\miniscule2},below of =A] (B) {\small$\texttt{b}$};
          \node[label={[xshift=0.4cm, yshift=-0.9cm]:\miniscule3},right of =A] (C) {\small$\texttt{s:c}$};
      \end{scope}

      \begin{scope}[>={Stealth[black]},
                    every node/.style={fill=white,rectangle,inner sep=0,fill=none},
                    every edge/.style={draw,thick}]
                      \path [->, -latex] (A) edge node[above=0.5pt,inner sep=1pt] {\small$\texttt{d}$} (C);
      \end{scope}
      \end{tikzpicture}
 };
\end{scope}
  \draw[vecArrow] (K) to (R);

  \draw[innerWhite] (K) to (R);
\end{tikzpicture}
\leftskip=0pt\small$\texttt{where s != "OUT"}$

\bigbreak
\leftskip=0pt\small$\texttt{unmark(a:list)}$
\medbreak
\centering
\begin{tikzpicture}
\begin{scope}[node distance=2.5cm, every node/.style={rectangle,minimum size=0.6cm,thick,draw=white,text=black}]
    \node (K) {
      \begin{tikzpicture}
      \begin{scope}[node distance=1.1cm, every node/.style={thick,draw}]
          \node[label={[xshift=0.4cm, yshift=-0.9cm]:\miniscule1},fill = nodeblue] (A) {\small$\texttt{a}$};
      \end{scope}
      \end{tikzpicture}
 };
 
    \node[right of=K] (R){
      \begin{tikzpicture}
      \begin{scope}[node distance=1.1cm, every node/.style={thick,draw}]
          \node[label={[xshift=0.4cm, yshift=-0.9cm]:\miniscule1}] (A) {\small$\texttt{a}$};
      \end{scope}
      \end{tikzpicture}
 };
\end{scope}
  \draw[vecArrow] (K) to (R);

  \draw[innerWhite] (K) to (R);
\end{tikzpicture}
\end{tabular}
\end{tabular}
\end{center}
\caption{A P-GP\,2 edge mutation $\texttt{MutateEdge}$ for digital circuits. This edge mutation preserves acylicity. The rule $\texttt{pick\_edge}$ is used to probabilistically choose an edge to mutate. Then $\texttt{mark\_output}$ is applied as long as possible, marking every node with a path to the source of the edge we wish to mutate blue. $\texttt{mutate\_edge}$ can then be applied safely, redirecting the edge to target some unmarked node which does not have a path to the source of the mutating edge. Finally \texttt{unmark} is applied as long as possible to return the graph to an unmarked state.\label{fig:dac_mutate}}
\end{figure*}
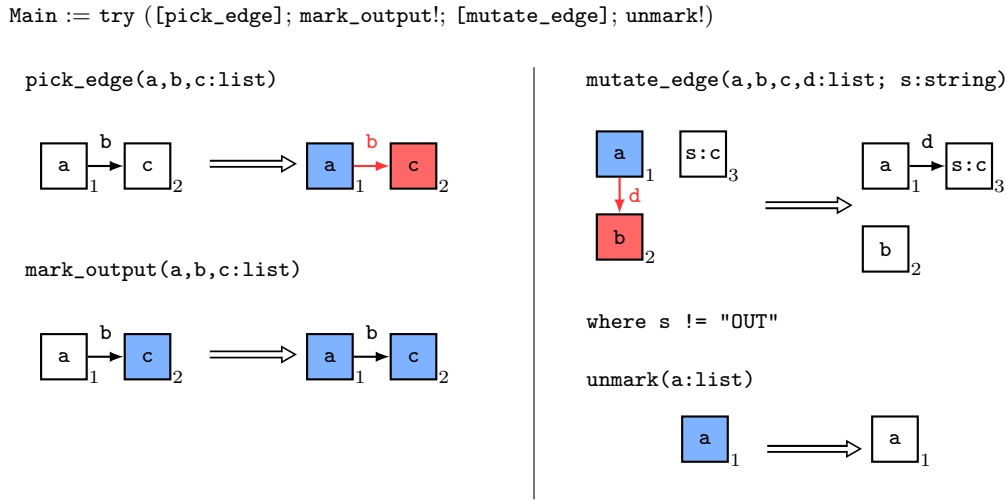

We directly encode digital circuits as graphs such that the graph contains input and output nodes (corresponding to the inputs and outputs of the intended problem) and function nodes.  In P-GP\,2, we identify input nodes and output nodes by labels of the form $\texttt{"IN"}:x$ and $\texttt{"OUT"}:y$ respectively, where $x$ and $y$ are integers that identify which particular input or output the node corresponds to. Function nodes are labelled as $\texttt{"$[f_i]$":a}$, where $[f_i]$ is a string uniquely identifying function $f_i \in F$ and $a$ is the arity of $f_i$. In this work our functions are symmetrical, but an extension is available to associate each edge with an integer to identify which particular input of a function it corresponds to. Fig. \ref{fig:dac_individual} shows a digital circuit encoded in this form.

For a specific $i$ input, $o$ output problem over function set $F$, we must evolve graphs that are constrained:
\begin{itemize}
\item Individual solutions are acyclic.
\item Individual solutions  have $i$ input nodes.
\item Individual solutions have $o$ output nodes.
\item All other nodes that are neither inputs nor outputs must be function nodes associated with some function $f_i \in F$ and have exactly $a$ outgoing edges where $a$ is the arity of $f_i$. 
\end{itemize}

We use three graph programs to induce a landscape; $\texttt{InitCircuit}$, $\texttt{MutateFunction}$ and $\texttt{MutateEdge}$. The first is the initialisation program  for generating individual graphs, and the others are mutation operators. $\texttt{InitCircuit}$ and $\texttt{MutateFunction}$ are given in Appendix \ref{appendix}; it should be clear that they satisfy the constraints described above. Here we describe in more detail the $\texttt{MutateEdge}$ operator, which is the mutation operator primarily responsible for the topological changes to individual solutions. 

\begin{figure}[!t]
\input{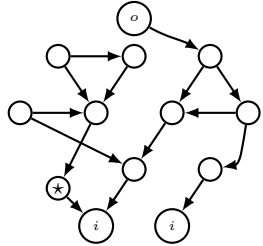}
\caption{A step-by-step execution of the edge mutation operator given in Figure \ref{fig:dac_mutate}. For visual simplicity, node labels have been omitted. \label{fig:viz_edge_mutate}}
\end{figure}

The $\texttt{MutateEdge}$ operator is shown in Fig. \ref{fig:dac_mutate}. It works by first picking an edge to mutate at random using the $\texttt{pick\_edge}$ rule, marking that edge red, its source blue and its target red. Then $\texttt{mark\_output}$ is applied as long as possible, marking blue every node for which there is a directed path to the source of the edge we wish to mutate. $\texttt{mutate\_edge}$ can be safely applied to redirect the edge to target some unmarked node (chosen at random); this cannot introduce a cycle as the new target is unmarked and therefore does not have a directed path to the existing source of the mutating edge. Finally $\texttt{unmark}$ is applied as long as possible to return the graph to an unmarked state. This P-GP\,2 program uses a uniform random distribution to chose the edge to mutate, a uniform distribution over all possible edge mutations that preserve acyclicity, and clearly respects the other constraints mentioned above, as it does not relabel any nodes or change the number of outgoing edges of any node. In \cite{Atkinson-Plump-Stepney18a} we argue that this edge mutation generalises the order preserving mutations of CGP and offers additional possible mutations. A visual step-by-step execution of this mutation operator is shown in Figure \ref{fig:viz_edge_mutate}.

In general, we use the $1+\lambda$ evolutionary algorithm with EGGP. $1+\lambda$ has been used extensively with CGP with favourable comparisons with large-population GP systems (see \cite{Miller2011-CGP}). A comparative study of crossover in CGP \cite{kalkreuth2018comparative} found that there is no currently known universal crossover operator for CGP and that $1+\lambda$ is sometimes the best known approach for certain problems. Current advice \cite{Miller2011-CGP,turner2015introducing} is to use $1+\lambda$ as the `standard' CGP approach. The comparative study between EGGP and CGP \cite{Atkinson-Plump-Stepney18a} exclusively used the $1+\lambda$ strategy with EGGP performing favourably on many digital circuit benchmark problems. In combination, these points appear to justify the exclusive use of $1+\lambda$ with EGGP in our study. Additionally, the use of $1+\lambda$ has the added effect of `isolating' our notion of semantic neutral drift, in that we can apply logical equivalence laws to the single surviving individual in each generation knowing that its application is not disrupting other processes e.g. crossover or non-elitist selection. 

\section{Semantic Neutral Drift \label{sec:NeutralEGGP}}

\subsection{The Concept \label{sec:snd1}}

Semantic Neutral Drift (SND) is the augmentation of a GP system with semantics-preserving mutations. These mutations are added to the standard mutation and cross\-over operators, which are intended to introduce variation to search. In this section we refer to mutation operators and individuals generally, not just our specific operation. For individual solutions $i,j$ and mutation operator $m$, we write $i \rightarrow_m j$ to mean that $j$ can be generated from $i$ by using mutation $m$.  A semantics-preserving mutation is one that guarantees that the semantic meaning of a child generated by that mutation is identical to that of its parent, for any choice of parents and a given semantic model. This definition is adequate for our domain of GP, where there is no distinction between the genotype and phenotype.

For our digital circuits case study, this semantic equivalence is well-defined: two circuits are semantically equivalent if they describe identical truth tables. Therefore, semantics preserving mutations in this context are ones which preserve an individual's truth table. As we will be evaluating individuals by the number of incorrect bits in their truth tables, there may be individuals with equivalent fitness but different truth tables. Therefore, semantic equivalence is distinct from, but related to, fitness equivalence. 

Additionally, semantics preserving mutations do not necessarily induce neutral drift. In the circumstance that a fitness function considers more than the semantics of an individual, there is no guarantee that the child of a parent generated by a semantics-preserving mutation has equal fitness to its parent. For example, if a fitness function penalized the size of an individual, a semantics-preserving mutation which introduces additional material (e.g. increases size) would generate children less fit than their parents under this measure. 

We identify a special class of fitness functions, where fitness depends only on semantics, and so where seman\-tics-preserving mutations are guaranteed to preserve fitness. In this circumstance, any use of semantics-pre\-ser\-ving mutations is a deliberate, designed-in, form of neutral drift. The fitness function in our case study is an example of this; the fitness of an individual depends only on its truth table. Formally we have the following: a set of semantics-preserving mutation operators $M$ over search space $S$ with respect to a fitness function $f$ that considers only semantics guarantees that

\[\forall i, j \in S, m \in M: (j \rightarrow_{m} i) \Rightarrow (f(i) = f(j)). \]

Consider a GP run that has reached a local optimum; no available mutations or crossover operators offer positive improvements with respect to the fitness function. It may be the case that there is a solution elsewhere in the landscape that is equally fit as the best found solution but has a neighborhood with positive mutations available. By applying a semantics preserving mutation to transform the best found solution into this other, semantically equivalent, solution, the evolutionary process gains access to this better neighborhood and can continue its search. Hence the proposed benefit of Semantic Neutral Drift is the same as conventional neutral drift: that by transforming discovered solutions we gain access to different parts of the landscape that may allow the population to escape local optima. The distinction here is that we are employing domain knowledge to deliberately preserve semantics, rather than accessing neutral drift as a byproduct of other evolutionary processes. The hypothesis we are investigating is that this deployment of domain knowledge yields more meaningful neutral mutations than simple rewrites of intronic code, and that this leads the evolutionary algorithm to more varied (and therefore useful) neighborhoods. 

\begin{figure}[tb]
\centering
\includegraphics[width=0.5\textwidth]{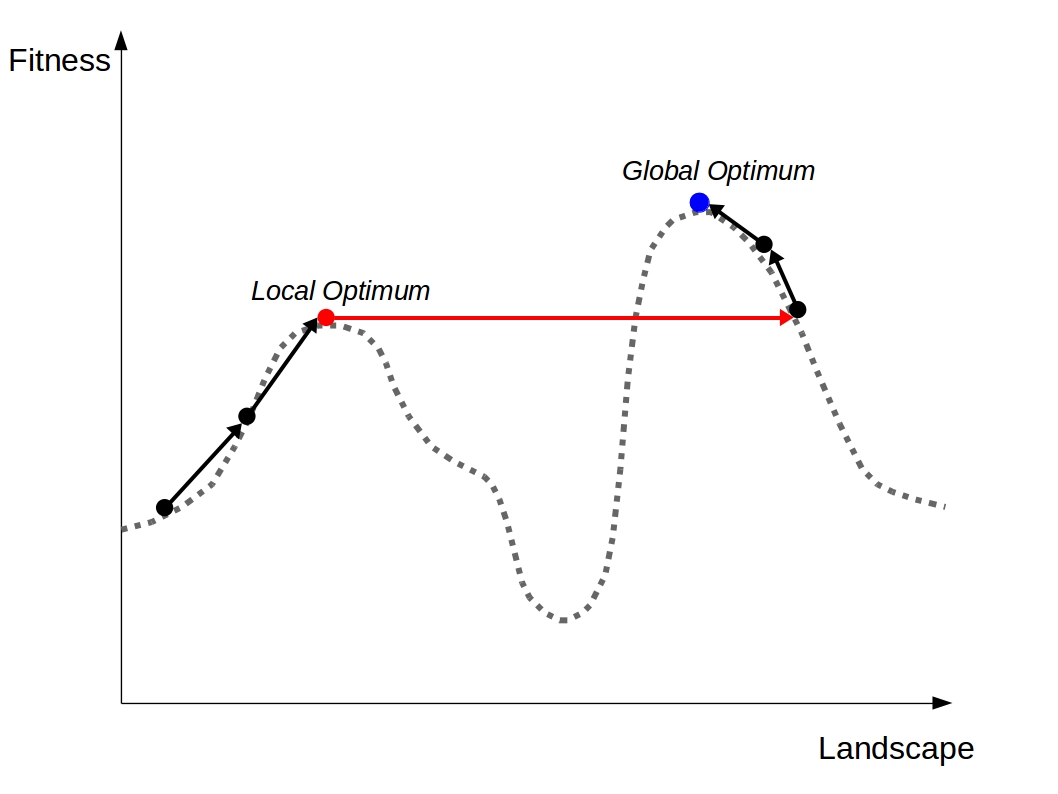}
\caption{\label{fig:snd}A simple visualization of Semantic Neutral Drift. Individuals exist in one dimension along the $x$-axis. On the $y$-axis, each individual has an associated fitness. Normal mutations (black arrows) allow the evolutionary algorithm to hill-climb by sampling from adjacent points. A semantics-preserving mutation (red arrow) allows the EA to leave a local optimum to move to a different slope where it can then climb to the global optimum.}
\end{figure}

A simple visualization of Semantic Neutral Drift is given in Figure \ref{fig:snd}. Here the landscape exists in one dimension (the $x$-axis) with fitness of individuals given in the $y$-axis. In this illustration, the individual has eached a local optimum, then a semantics-preserving mutation moves it to a different `hill' from which it is able to reach the global optimum. 

While our experiments will focus on the role of semantic neutral drift when evolving graphs with EGGP, we argue that the underlying concept is extendable to other GP systems. For example, Koza noted the possibility of applying DeMorgan's laws to GP trees \cite[Ch.6]{koza:book} which, if used in a continuous process rather than as a solution optimiser, would induce semantic neutral drift. It is also plausible to apply similar operators to CGP \cite{Miller2000-CGP} representations, although the ordering imposed on the representation raises some technical difficulties with respect to where newly created nodes should be placed. The potential for Embedded CGP \cite{Walker2008} to effectively grow and shrink the overall size of the genotype offers some hope in this direction.


\subsection{Designing Semantic Neutral Drift}

We extend EGGP by applying semantics-preserving mutations to members of the population each generation. Here we focus on digital circuits as a case study, and design mutations which modify the \textit{active} components of the individual by exploiting domain knowledge of logical equivalence.

For the function set $\{\texttt{AND,OR,NOT}\}$ there are a number of known logical equivalences. Here we use DeMorgan's laws:
\begin{center}
$\mbox{DeMorgan$_{F1}$: }\neg(a \land b) = \neg a \lor \neg b$\\
$\mbox{DeMorgan$_{F2}$: }\neg(a \lor b) = \neg a \land \neg b$\\
$\mbox{DeMorgan$_{R1}$: }\neg a \lor \neg b = \neg(a \land b)$\\
$\mbox{DeMorgan$_{R2}$: }\neg a \land \neg b = \neg(a \lor b)$
\end{center}
and the identity and double negation laws:
\begin{center}
$\mbox{ID-AND$_{F}$: } a = a \land a$\\
$\mbox{ID-AND$_{R}$: } a \land a = a$\\
$\mbox{ID-OR$_{F}$: } a = a \lor a$\\
$\mbox{ID-OR$_{R}$: } a \lor a = a$\\
$\mbox{ID-NOT$_{F}$: } a = \neg\neg a$\\
$\mbox{ID-NOT$_{R}$: } \neg\neg a = a$
\end{center}
Here we investigate different subsets of these semantics-preserving rules. We encode them as graph transformation rules to apply to the active component of an individual. In the context of the $1+\lambda$ evolutionary algorithm, we apply one of the rules from the subset to the surviving individual of each generation. 

\begin{figure*}
\input{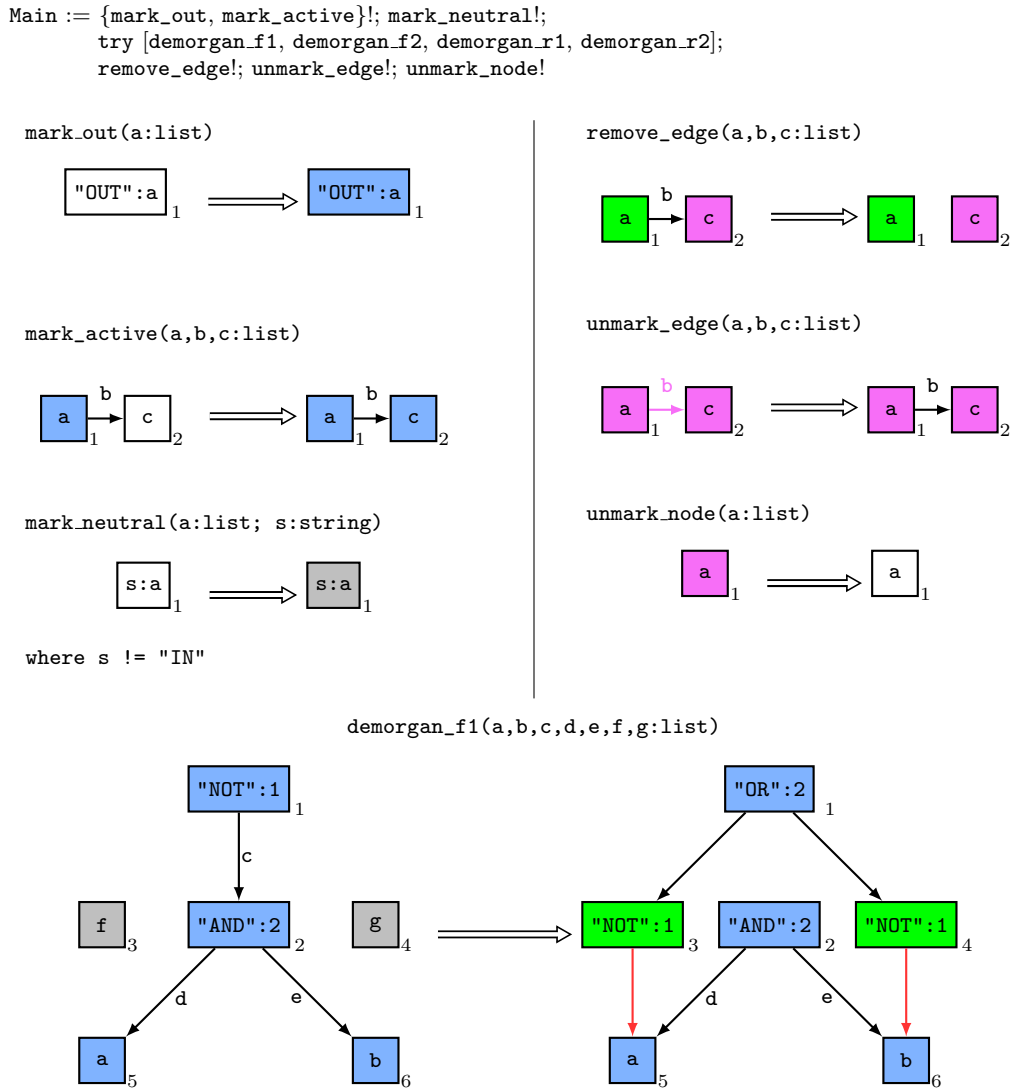}
\caption{A P-GP\,2 program for performing semantics preserving mutations to digital circuits.\label{fig:seman}}
\end{figure*}

Encoding these semantics-preserving rules is non-trivial for our individuals as they incorporate sharing; multiple nodes may use the same node as an input, and therefore rewriting or removing that node, e.g. as part of DeMorgan's, may disrupt the semantics elsewhere in the individual. To overcome this, we need a more sophisticated rewriting program.  The graph program in Fig. \ref{fig:seman} is designed for the logical equivalence laws DeMorgan$_{F1|F2}$, 
DeMorgan$_{R1|R2}$; 
 analogous programs are used for other operators. 

The program $\texttt{Main}$ in Fig. \ref{fig:seman} works as follows. 

\{\texttt{mark\_out}, \texttt{mark\_active}\}! :
Mark all active nodes with the given rule-set  applied as long as possible. Once this rule-set has no matches, all inactive nodes must be unmarked: these are `neutral' nodes that do not contribute to the semantics of the individual.

\texttt{mark\_neutral}! :
Mark these neutral nodes grey with the rule  applied as long as possible. We can then rewrite the individual while preserving semantics with respect to shared nodes by incorporating neutral nodes into the active component rather than overwriting existing nodes. 

\texttt{try} [\texttt{demorgan\_f1}, \texttt{demorgan\_f2}, \texttt{demorgan\_r1}, \linebreak \texttt{demorgan\_r2}] :
pick some rule with uniform probability from the subset of the listed rules that have valid matches. When a rule has been chosen, a match is chosen for it from the set of all possible matches with uniform probability. The probabilistic rule-set call is surrounded by a $\texttt{try}$ statement to catch the fail case that none of the rules have matches. 

In Fig. \ref{fig:seman} we show one of the 4 referenced rules, $\texttt{demorgan$\_$f1}$, which corresponds to the logical equivalence law DeMorgan$_{F1}$; the others may be given analogously. On the left hand side is a match for the pattern $\neg(a \land b)$ in the active component and 2 neutral nodes. If the matched pattern were directly transformed, any nodes sharing use of the matches for node $2$ or node $3$ could have their semantics disrupted. Instead, the right-hand-side of $\texttt{demorgan$\_$f1}$ changes the syntax of node $1$ to correspond to $\neg a \lor \neg b$ by absorbing the matched neutral nodes (preserving its semantics) without rewriting nodes $1$ or $2$ and disrupting their semantics. Nodes $3$ and $4$ are marked green and their newly created outgoing edges are marked red. These marks are used later in the program to clean up any previously existing outgoing edges they have to other parts of the graph.

\texttt{remove$\_$edge}:
once a semantics preserving rule has been applied, the rule  is applied as long as possible to remove the other outgoing edges of green marked absorbed nodes. 

\texttt{unmark$\_$edge!; unmark$\_$node!}:
return the graph to an unmarked state, where nodes and edges with any mark (indicated by magenta edges and nodes in the rules) have their marks removed. 

This program highlights the helpfulness of graph programming for this task. The probabilistic application of complex transformations, such as DeMorgan's law, to only the active components of a graph-like program with sharing is non-trivial, but can be concisely described by a graph program.

\subsection{Variations on our approach}

We identify 3 sets of logical equivalence rules to study, alongside another example of semantics preserving transformation taken from term-rewriting theory. These sets are detailed in Table \ref{tab:semset}. The first 3 sets comprise the logical equivalence laws already discussed. The last, CC, refers to collapsing and copying from term graph rewriting (see \cite{HKP88}). Collapsing is the process of merging semantically equivalent subgraphs, and copying is the process of duplicating a subgraph. 

\begin{figure*}
\hspace{3.0cm}\small$\texttt{copy\char`_2(a,b,c,d,e,f,g,h,i,j:list; s:string)}$\quad
\medbreak
\centering
\begin{tikzpicture}
\begin{scope}[node distance=5cm, every node/.style={rectangle,minimum size=0.6cm,thick,draw=white,text=black}]
    \node (K) {
      \begin{tikzpicture}
      \begin{scope}[node distance=1.8cm, every node/.style={thick,draw}]
          \node[label={[xshift=0.4cm, yshift=-0.9cm]:\miniscule1},fill = nodeblue] (A) {\small$\texttt{a}$};
          \node[label={[xshift=0.5cm, yshift=-0.9cm]:\miniscule2},below of =A,fill = nodeblue] (B) {\small$\texttt{s:2}$};
          \node[label={[xshift=0.4cm, yshift=-0.9cm]:\miniscule3},right of = A,fill = nodeblue] (C) {\small$\texttt{b}$};
          \node[label={[xshift=0.4cm, yshift=-0.9cm]:\miniscule4},right of =B,fill = lightgray] (D) {\small$\texttt{j}$};
          \node[label={[xshift=0.4cm, yshift=-0.9cm]:\miniscule5},below of =B,fill = nodeblue] (E) {\small$\texttt{h}$};
          \node[label={[xshift=0.4cm, yshift=-0.9cm]:\miniscule6},below of =D,fill = nodeblue] (F) {\small$\texttt{i}$};
      \end{scope}

      \begin{scope}[>={Stealth[black]},
                    every node/.style={fill=white,rectangle,inner sep=0,fill=none},
                    every edge/.style={draw,thick}]
                      \path [->, -latex] (A) edge node[xshift=3pt,inner sep=1pt] {\small$\texttt{c}$} (B);
                      \path [->, -latex] (B) edge node[xshift=4pt,yshift=-2pt,inner sep=1pt] {\small$\texttt{f}$} (E);
                      \path [->, -latex] (B) edge node[xshift=9pt,yshift=-3pt,inner sep=1pt] {\small$\texttt{g}$} (F);
                      \path [->, -latex] (C) edge node[xshift=4pt,yshift=-1pt,inner sep=1pt] {\small$\texttt{d}$} (B);
      \end{scope}
      \end{tikzpicture}
 };
 
    \node[right of=K] (R){
      \begin{tikzpicture}
      \begin{scope}[node distance=1.8cm, every node/.style={thick,draw}]
          \node[label={[xshift=0.4cm, yshift=-0.9cm]:\miniscule1},fill = nodeblue] (A) {\small$\texttt{a}$};
          \node[label={[xshift=0.5cm, yshift=-0.9cm]:\miniscule2},below of =A,fill = nodeblue] (B) {\small$\texttt{s:2}$};
          \node[label={[xshift=0.4cm, yshift=-0.9cm]:\miniscule3},right of = A,fill = nodeblue] (C) {\small$\texttt{b}$};
          \node[label={[xshift=0.5cm, yshift=-0.9cm]:\miniscule4},right of =B,fill = green] (D) {\small$\texttt{s:2}$};
          \node[label={[xshift=0.4cm, yshift=-0.9cm]:\miniscule5},below of =B,fill = nodeblue] (E) {\small$\texttt{h}$};
          \node[label={[xshift=0.4cm, yshift=-0.9cm]:\miniscule6},below of =D,fill = nodeblue] (F) {\small$\texttt{i}$};
      \end{scope}

      \begin{scope}[>={Stealth[black]},
                    every node/.style={fill=white,rectangle,inner sep=0,fill=none},
                    every edge/.style={draw,thick}]
                      \path [->, -latex] (A) edge node[xshift=3pt,inner sep=1pt] {\small$\texttt{c}$} (B);
                      \path [->, -latex] (B) edge node[xshift=4pt,yshift=-2pt,inner sep=1pt] {\small$\texttt{f}$} (E);
                      \path [->, -latex] (B) edge node[xshift=9pt,yshift=-3pt,inner sep=1pt] {\small$\texttt{g}$} (F);
                      \path [->, -latex] (C) edge node[xshift=4pt,yshift=-1pt,inner sep=1pt] {\small$\texttt{d}$} (D);
                      \path [->, -latex, edgered, draw=edgered] (D) edge (F);
                      \path [->, -latex, edgered, draw=edgered] (D) edge (E);
      \end{scope}
      \end{tikzpicture}
 };
\end{scope}
  \draw[vecArrow] (K) to (R);

  \draw[innerWhite] (K) to (R);
\end{tikzpicture} 

\leftskip=0pt\hspace{3.0cm}\small$\texttt{collapse\char`_2(a,b,c,d,e,f,g,h,i,j:list; s:string)}$\quad
\medbreak
\centering
\begin{tikzpicture}
\begin{scope}[node distance=5cm, every node/.style={rectangle,minimum size=0.6cm,thick,draw=white,text=black}]
    \node (K) {
      \begin{tikzpicture}
      \begin{scope}[node distance=1.8cm, every node/.style={thick,draw}]
          \node[label={[xshift=0.4cm, yshift=-0.9cm]:\miniscule1},fill = nodeblue] (A) {\small$\texttt{a}$};
          \node[label={[xshift=0.5cm, yshift=-0.9cm]:\miniscule2},below of =A,fill = nodeblue] (B) {\small$\texttt{s:2}$};
          \node[label={[xshift=0.4cm, yshift=-0.9cm]:\miniscule3},right of = A,fill = nodeblue] (C) {\small$\texttt{b}$};
          \node[label={[xshift=0.5cm, yshift=-0.9cm]:\miniscule4},right of =B,fill = nodeblue] (D) {\small$\texttt{s:2}$};
          \node[label={[xshift=0.4cm, yshift=-0.9cm]:\miniscule5},below of =B,fill = nodeblue] (E) {\small$\texttt{i}$};
          \node[label={[xshift=0.4cm, yshift=-0.9cm]:\miniscule6},below of =D,fill = nodeblue] (F) {\small$\texttt{j}$};
      \end{scope}

      \begin{scope}[>={Stealth[black]},
                    every node/.style={fill=white,rectangle,inner sep=0,fill=none},
                    every edge/.style={draw,thick}]
                      \path [->, -latex] (A) edge node[xshift=3pt,inner sep=1pt] {\small$\texttt{c}$} (B);
                      \path [->, -latex] (B) edge node[xshift=4pt,yshift=-2pt,inner sep=1pt] {\small$\texttt{e}$} (E);
                      \path [->, -latex] (B) edge node[xshift=9pt,yshift=-2pt,inner sep=1pt] {\small$\texttt{g}$} (F);
                      \path [->, -latex] (C) edge node[xshift=4pt,yshift=-1pt,inner sep=1pt] {\small$\texttt{d}$} (D);
                      \path [->, -latex] (D) edge node[xshift=-9pt,yshift=-2pt,inner sep=1pt] {\small$\texttt{f}$} (E);
                      \path [->, -latex] (D) edge node[xshift=4pt,yshift=-1pt,inner sep=1pt] {\small$\texttt{h}$} (F);
      \end{scope}
      \end{tikzpicture}
 };
 
    \node[right of=K] (R){
      \begin{tikzpicture}
      \begin{scope}[node distance=1.8cm, every node/.style={thick,draw}]
          \node[label={[xshift=0.4cm, yshift=-0.9cm]:\miniscule1},fill = nodeblue] (A) {\small$\texttt{a}$};
          \node[label={[xshift=0.5cm, yshift=-0.9cm]:\miniscule2},below of =A,fill = nodeblue] (B) {\small$\texttt{s:2}$};
          \node[label={[xshift=0.4cm, yshift=-0.9cm]:\miniscule3},right of = A,fill = nodeblue] (C) {\small$\texttt{b}$};
          \node[label={[xshift=0.5cm, yshift=-0.9cm]:\miniscule4},right of =B,fill = nodeblue] (D) {\small$\texttt{s:2}$};
          \node[label={[xshift=0.4cm, yshift=-0.9cm]:\miniscule5},below of =B,fill = nodeblue] (E) {\small$\texttt{i}$};
          \node[label={[xshift=0.4cm, yshift=-0.9cm]:\miniscule6},below of =D,fill = nodeblue] (F) {\small$\texttt{j}$};
      \end{scope}

      \begin{scope}[>={Stealth[black]},
                    every node/.style={fill=white,rectangle,inner sep=0,fill=none},
                    every edge/.style={draw,thick}]
                      \path [->, -latex] (A) edge node[xshift=3pt,inner sep=1pt] {\small$\texttt{c}$} (B);
                      \path [->, -latex] (B) edge node[xshift=4pt,yshift=-2pt,inner sep=1pt] {\small$\texttt{e}$} (E);
                      \path [->, -latex] (B) edge node[xshift=9pt,yshift=-2pt,inner sep=1pt] {\small$\texttt{g}$} (F);
                      \path [->, -latex] (C) edge node[xshift=4pt,yshift=-1pt,inner sep=1pt] {\small$\texttt{d}$} (B);
                      \path [->, -latex] (D) edge node[xshift=-9pt,yshift=-2pt,inner sep=1pt] {\small$\texttt{f}$} (E);
                      \path [->, -latex] (D) edge node[xshift=4pt,yshift=-1pt,inner sep=1pt] {\small$\texttt{h}$} (F);
      \end{scope}
      \end{tikzpicture}
 };
\end{scope}
  \draw[vecArrow] (K) to (R);

  \draw[innerWhite] (K) to (R);
\end{tikzpicture} 
\caption{The rules $\texttt{copy\_2}$ and $\texttt{collapse\_2}$. The rule $\texttt{copy\_2}$ matches a 2-arity function node that is shared by 2 active nodes and absorbs a neutral node to effectively copy that 2-arity function node and redirect one of the original node's shared incoming edges to that copy. The rule $\texttt{collapse\_2}$ attempts the reverse of $\texttt{copy\_2}$ by matching 2 active identical 2-arity function nodes and redirecting one of those nodes' incoming edges to the other. The node which has lost an incoming edge, if it was shared by no other nodes, may now become neutral.\label{fig:cc}}
\end{figure*}
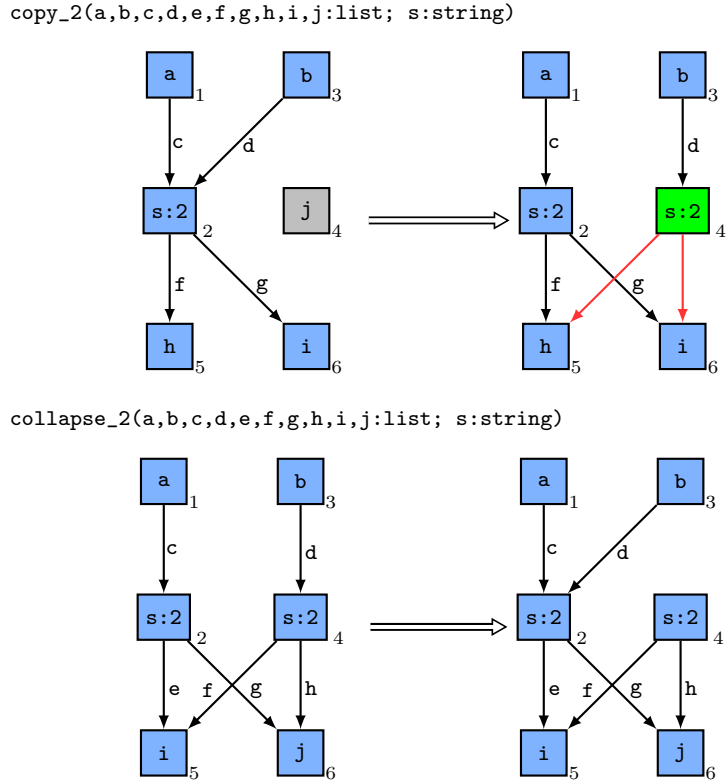

The rules $\texttt{collapse}_2$ and $\texttt{copy}_2$ are shown in Fig.~\ref{fig:cc}. These collapse and copy, respectively, function nodes of arity 2 without garbage collection. We only require rules for arity 1 and arity 2 as our function sets in experiments are limited to arity 2. This final set is included for several reasons: it takes a different form from the domain-specific logical equivalence laws in the other 3 sets; it allows us to investigate if the apparent overlap between term-graph rewriting and evolutionary algorithms bears fruit; it appears to resemble gene duplication, which is a natural biological process believed to aid evolution \cite{zhang2003evolution}.

\begin{table}[tp]
\centering
\begin{tabular}{C{2.9cm} C{4.7cm}}
\toprule
\textbf{Set} &
\textbf{Rules} \\
\midrule
DeMorgan (DM) & DeMorgan$_{F1}$, DeMorgan$_{F2}$, DeMorgan$_{R1}$, DeMorgan$_{R2}$ \\
\midrule
DeMorgan and Negation (DMN) & DeMorgan$_{F1}$, DeMorgan$_{F2}$, DeMorgan$_{R1}$, DeMorgan$_{R2}$, ID-NOT$_F$, ID-NOT$_R$ \\
\midrule
Identity (ID) & ID-AND$_F$, ID-AND$_R$, ID-OR$_F$, ID-OR$_R$, ID-NOT$_F$, ID-NOT$_R$ \\
\midrule
Collapse/Copy (CC) & collapse$_1$, collapse$_2$, copy$_1$, copy$_2$ \\
\bottomrule
\end{tabular}
\caption{The studied semantics preserving rule-sets.}
\label{tab:semset}
\end{table}

\section{Experimental Setup \label{sec:Exp}}

To evaluate our approach, we study the same digital circuit benchmark problems as in \cite{Atkinson-Plump-Stepney18a}, listed in Table~\ref{tab:dcpset}. We perform 100 runs of each of our 4 neutral drift sets (Table~\ref{tab:semset}) on each problem (Table~\ref{tab:dcpset}).  We use the $1+\lambda$ evolutionary algorithm with $\lambda=4$. We use a mutation rate of $0.01$ and fix all individuals to use $100$ function nodes.  The fitness function used is the number of incorrect bits in an individual's truth table compared to the target truth table, hence we are minimizing the fitness. We are able to achieve $100\%$ success rate in finding global optima in our evolutionary runs, so we compare the number of evaluations required to find perfect fitness.

The function set used here is $\{\texttt{AND}, \texttt{OR}, \texttt{NOT}\}$, rather than the set $\{\texttt{AND}, \texttt{OR}, \texttt{NAND}, \texttt{NOR}\}$  used in \cite{Atkinson-Plump-Stepney18a} and  \cite[Ch.2]{Miller2011-CGP}. Our function set is chosen to directly correspond to the logical equivalence laws used. To give context to the results in Section \ref{sec:Results}, and to highlight that the chosen function set is the harder of the two, we run EGGP with both function sets and detail the results in Table~\ref{tab:EGGPres}. For additional context, the comparative study in \cite{Atkinson-Plump-Stepney18a} has shown EGGP to perform favourably in comparison to CGP on these problems with the $\{\texttt{AND}, \texttt{OR}, \texttt{NAND}, \texttt{NOR}\}$ function set.

We use a two-tailed Mann--Whit\-ney $U$ test \cite{mann1947} to establish a statistically significant difference between the median number of evaluations using the two different function sets.
When a result is statistically significant ($p < 0.05$) we also use a Vargha--Delaney $A$ test \cite{vargha00} to measure the effect size.
On every problem, using \{\texttt{AND}, \texttt{OR}, \texttt{NOT}\} takes significantly ($p < 0.05$) more effort (in terms of evaluations) than when using \{\texttt{AND}, \texttt{OR}, \texttt{NAND}, \texttt{NOR}\}.
This justifies our assertion that the former function set is `harder' to evolve.

\begin{table}[tp]
\centering
\begin{tabular}{lR{1.4cm}R{1.4cm}}
\toprule
\textbf{Digital Circuit} &
\textbf{No. Inputs} &
\textbf{No. Outputs} \\
\midrule
1-bit Adder (1-Add)           & 3 & 2                          \\
2-bit Adder (2-Add)           & 5 & 3                          \\
3-bit Adder (3-Add)           & 7 & 4                          \\
2-bit Multiplier (2-Mul)           & 4 & 4                          \\ 
3-bit Multiplier (3-Mul)           & 6 & 6                          \\ 
3:8-bit De-Multiplexer (DeMux)           & 3 & 8                          \\ 
4$\times$1-bit Comparator (Comp)         & 4 & 18     \\                      
3-bit Even Parity (3-EP)         & 3 & 1                 \\     
4-bit Even Parity (4-EP)         & 4 & 1                    \\
5-bit Even Parity (5-EP)         & 5 & 1                           \\ 
6-bit Even Parity (6-EP)         & 6 & 1                           \\ 
7-bit Even Parity (7-EP)         & 7 & 1                           \\ 
\bottomrule
\end{tabular}
\caption{Digital Circuit benchmark problems.}
\label{tab:dcpset}
\end{table}

\begin{table}[tp]
\setlength{\tabcolsep}{2pt}
\centering
\begin{tabular}{l rrrrrcc}
\toprule
& \multicolumn{5}{c}{\textbf{EGGP}}\\
Problem& \multicolumn{2}{c}{$\{\texttt{AND}, \texttt{OR}, \texttt{NOT}\}$} & \, & \multicolumn{2}{c}{$\{\texttt{AND}, \texttt{OR}, \texttt{NAND}, \texttt{NOR}\}$} & \, & \,\\\cline{2-3}\cline{5-6}
& \multicolumn{1}{c}{ME} & \multicolumn{1}{c}{IQR} & & \multicolumn{1}{c}{ME} & \multicolumn{1}{c}{IQR} & $p$ & $A$ \\
\midrule
1-Add  & 15,538 & 18,963 & & 7,495 & 8,764 & $10^{-7}$ & 0.71\\
2-Add  & 162,003 & 172,781 & & 82,688  & 79,333 & $10^{-8}$ & \textbf{0.73}\\
3-Add  & 742,948 & 679,040 & & 309,570 & 288,865 & $10^{-16}$ & \textbf{0.83}\\
\midrule
2-Mul  & 21,733 & 28,319 & & 14,263 & 13,801 & $10^{-4}$ & 0.65\\
3-Mul  & 1,326,880 & 907,544 & & 932,430 & 643,529 & $10^{-6}$ & 0.68\\
\midrule
DeMux  & 28,123 & 17,450 & & 17,100 & 10,763 & $10^{-9}$ & \textbf{0.75}\\
Comp  & 408,448 & 275,581 & & 147,343 & 128,304 & $10^{-17}$ & \textbf{0.85}\\
\midrule
3-EP  & 7,403 & 8,051 & & 4,295 & 5,500 & $10^{-4}$ & 0.66\\
4-EP  & 26,715 & 20,430 & & 16,445 & 13,568 & $10^{-9}$ & \textbf{0.73}\\
5-EP  & 76,608 & 57,518 & & 42,778 & 29,454 & $10^{-10}$ & \textbf{0.75}\\
6-EP  & 175,908 & 120,504 & & 80,940 & 56,283 & $10^{-15}$ & \textbf{0.83}\\
7-EP  & 380,600 & 237,965 & & 157,755 & 118,065 & $10^{-19}$ & \textbf{0.87}\\
\bottomrule
\end{tabular}
\caption{Baseline results from Digital Circuit benchmarks for EGGP on the $\{\texttt{AND}, \texttt{OR}, \texttt{NOT}\}$ and $\{\texttt{AND}, \texttt{OR}, \texttt{NAND}, \texttt{NOR}\}$ function sets. ME/IQR: the median/inter-quartile range of the number of evaluations used to solve the problem. The $p$ value is from the two-tailed Mann-Whitney $U$ test. Where $p<0.05$, the effect size  from the Vargha-Delaney A test is shown; large effect sizes ($A>0.71$) are shown in \textbf{bold}.}
\label{tab:EGGPres}
\end{table}
\section{Results \label{sec:Results}}

\begin{table*}[tp]
\setlength{\tabcolsep}{2.5pt}
\centering
\begin{tabular}{l rccc rccc rccc rccc}
\toprule
& \multicolumn{15}{c}{\textbf{Neutral Ruleset}}\\
\textbf{Circuit}& \multicolumn{3}{c}{\textbf{DM}} & \, & \multicolumn{3}{c}{\textbf{DMN}} & \, & \multicolumn{3}{c}{\textbf{ID}} & \,& \multicolumn{3}{c}{\textbf{CC}} \\\cline{2-4}\cline{6-8} \cline{10-12} \cline{14-16}
& \multicolumn{1}{c}{ME} & $p$ & $A$ & & \multicolumn{1}{c}{ME} & $p$ & $A$ & & \multicolumn{1}{c}{ME} & $p$ & $A$ & & \multicolumn{1}{c}{ME} & $p$ & $A$ \\
\midrule
1-Add  & 8,950 & $10^{-7}$ & \textbf{0.72} & & 9,893 & $10^{-5}$ & 0.68 & & 9,093 & $10^{-7}$ & 0.71 & & 8,275 & $10^{-7}$ & \textbf{0.72}  \\
2-Add  & 65,692 & $10^{-14}$ & \textbf{0.81} &  & 49,200 & $10^{-21}$ & \textbf{0.88} & & 73,275 & $10^{-12}$ & \textbf{0.79} & & 103,393 & $10^{-5}$ & 0.68 \\
3-Add  & 255,003 & $10^{-19}$ & \textbf{0.87} &  & 186,647 & $10^{-25}$ & \textbf{0.93} & & 279,140 & $10^{-18} $& \textbf{0.86} & & 592,815 & 0.09 & -- \\
\midrule
2-Mul  & 19,853 & 0.36 & -- & & 16,680 & 0.01 & 0.60 & & 13,312 & $10^{-7}$ & 0.71 & & 19,995 & 0.29 & -- \\
3-Mul  & 955,418 & $10^{-3}$ & 0.63 &  & 678,403 & $10^{-11}$ & \textbf{0.77} & & 591,748 & $10^{-22}$ & \textbf{0.89} & & 975,558 & $10^{-4}$ & 0.65 \\
\midrule
DeMux  & 19,633 & $10^{-5}$ & 0.68 & & 16,678 & $10^{-12}$ & \textbf{0.79} & & 29,700 & 0.59 & -- & & 19,098 & $10^{-5}$ & 0.67 \\
Comp  & 542,290 & $10^{-3}$ & 0.63 & & 453,730 & 0.44 & -- & & 298,758 & $10^{-4}$ & 0.66 & & 576,263 & $10^{-4}$ & 0.64 \\
\midrule
3-EP  & 6,283 & 0.05 & -- & & 5,248 & $10^{-3}$ & 0.61 & & 5,990 & $10^{-3}$ & 0.61 & & 5,860 & 0.08 & -- \\
4-EP  & 23,828 & 0.06 & -- &  & 20,278 & $10^{-5}$ & 0.66 & & 18,745 & $10^{-6}$ & 0.69 & & 20,295 & $10^{-3}$ & 0.62        \\
5-EP  & 57,333 & 0.01 & 0.60 & & 58,408 & $10^{-3}$ & 0.62 & & 43,313 & $10^{-10}$ & \textbf{0.76} & & 60,087 & 0.01  & 0.60 \\
6-EP  & 129,910 & $10^{-5}$ & 0.67 & & 134,770 & 0.03 & 0.58 & & 104,392 & $10^{-9}$ & \textbf{0.74} & & 113,037 & $10^{-6}$ & 0.68 \\
7-EP  & 232,735 & $10^{-9}$ & \textbf{0.75} & & 330,572 & 0.05 & 0.58 & & 221,790 & $10^{-12}$ & \textbf{0.78} & & 219,237 & $10^{-12}$ & \textbf{0.78} \\
\bottomrule
\end{tabular}
\caption{Results from Digital Circuit benchmarks for the various proposed neutral rule-sets. The $p$ value is from the two-tailed Mann-Whitney $U$ test. Where $p<0.05$, the effect size  from the Vargha-Delaney A test is shown; large effect sizes ($A>0.71$) are shown in \textbf{bold}.}
\label{tab:digres}
\end{table*}

The results from our experiments are given in Table~\ref{tab:digres}. Each neutral rule-set is listed with the median evaluations (ME) required to solve each benchmark problem. 

We use a two-tailed Mann--Whitney $U$ test to demonstrate statistical significance in the difference of the median evaluations for these runs and the unmodified EGGP results given in Table \ref{tab:EGGPres}.

\begin{table*}[t]
\setlength{\tabcolsep}{2pt}
\centering
\begin{tabular}{lrrrrrrrrrccccc}
\toprule
Problem& \multicolumn{2}{c}{DMN} & \, & \multicolumn{2}{c}{ID} & \, & \multicolumn{2}{c}{EGGP} & \,&   \multicolumn{5}{c}{$p$}\\\cline{2-3}\cline{5-6}\cline{8-9}
& \multicolumn{1}{c}{MAS} & \multicolumn{1}{c}{IQR} & & \multicolumn{1}{c}{MAS} & \multicolumn{1}{c}{IQR} &  & \multicolumn{1}{c}{MAS} & \multicolumn{1}{c}{IQR} & \, & DMN vs. ID & \, & DMN vs. EGGP & \, & ID vs. EGGP\\
\midrule
3-Add  & 96.9 & 1.3 & & 92.3 & 1.2 & & 50.8 & 2.6 & & $10^{-33}$ && $10^{-34}$ && $10^{-34}$\\
Comp  & 99.3 & 95.6 & & 92.3 & 0.5 & & 67.0 & 2.3 & & $10^{-34}$ && $10^{-34}$ && $10^{-34}$\\
\bottomrule
\end{tabular}
\caption{Observed average solution size of the surviving population for the DMN rule-set, ID rule-set and EGGP without a neutral rule-set. Results are for the 3-Bit Adder (3-Add) and 4$\times$1-Bit Comparator (Comp) problems. For each result, the Median Average Size (MAS) and Interquartile Range (IQR) are given. The $p$ value is from the two-tailed Mann-Whitney $U$ test.}
\label{tab:Sizeres}
\end{table*}

For most problems and neutral rule-sets, the inclusion of semantic neutral drift yields statistically significant improvements in performance. There are some exceptions: for the 4$\times$1-bit comparator (COMP) problem, the inclusion of neutral rule-sets leads either to insignificant differences or to significantly worse performance for every rule-set except ID, which performs significantly better. The DeMorgan's rule-set (DM) and Copy/Collapse rule-set (CC) appear to yield the smallest benefit, finding significant improvement on only 8 and 9 of the 13 benchmark problems respectively. Additionally, both of these rule-sets yield significantly worse performance for the 4$\times$1-bit comparator (COMP) problem. The DeMorgan's and Negation rule-set (DMN) offer the best performance on the 2-bit and 3-bit adder problems (2-Add and 3-Add), in terms of median evaluations, $p$ value and effect size. The Identity rule-set (ID) achieves the best performance on the 2-bit and 3-bit multiplier problems (2-Mul and 3-Mul) but fails to achieve significant improvements on the 3:8-bit de-multiplexer problem (DeMux).

Our results show that, for some problems and certain neutral rule-sets, the inclusion of neutral drift may improve performance with respect to the effort (measured by the number of evaluations) required. Additionally, they offer strong evidence for the claim that there are some neutral rule-sets which may generally improve performance for a wide range of problems, particularly evidenced by the DMN and ID rule-sets. 

We identify DMN and ID as the best performing rule-sets; each of these yield significant improvements in performance across all but one problems (the exceptions being Comp and DeMux, respectively), and on those single problems that they fail to improve upon, their inclusion does not lead to significant detriment in performance. For this reason, these rule-sets are the subject of further analysis in Section \ref{sec:Analysis}.

\section{Analysis \label{sec:Analysis}}

\subsection{Neutral Drift or Neutral Growth?}

Analysis of the runtime of EGGP augmented with the DMN and ID neutral rule-sets reveals their bias towards searching the space of larger solutions. When we refer to larger solutions, given that EGGP uses fixed-size representations, we refer to the proportion of the individual graph which is active, defined by the number of nodes to which there is a path from an output node. We demonstrate this with the results given in Table \ref{tab:Sizeres}. Here, we measure the average (mean) size of the single surviving member throughout evolutionary runs on the 3-Add and Comp problems and give the median and interquartile range of these average sizes over 100 runs. The size of an individual is the number of active function nodes (those which are reachable from output nodes) contained within it. We give these values for DMN, ID and EGGP alone. We use a two-tailed Mann-Whitney $U$ test to measure for statistical differences between these observations. On both problems, DMN has a higher median average size (MAS) than both ID and EGGP alone ($p < 0.05$) and ID also has a higher MAS than EGGP alone ($p < 0.05$).

This observation challenges existing ideas that increasing the proportion of inactive code aids evolution \cite{miller2006redundancy}. We are able to achieve improvements in performance while effectively reducing the proportion of inactive code. It may be the case that high proportions of inactive code are helpful only when other forms of neutral drift are not available. 

The result that DMN and ID increase the active size of individuals initially appears to challenge our hypothesis that it is semantic neutral drift that aids evolution. An alternative explanation could be that it is `neutral growth', where our neutral rule-sets increase the size of individuals, that biases search towards larger solutions, which then happen to be better candidates for the problems we study. However, the CC neutral rule-set exclusively features neutral growth and neutral shrinkage, exploiting no domain knowledge beyond the notion that identical nodes in identical circumstances perform the same functionality, and featuring no meaningful semantic rewriting. We  therefore compare how CC and DMN perform with different numbers of nodes available, to determine whether larger solutions are indeed better candidates for the studied problems.

We run DMN, CC and standard EGGP on the 2-Add, 3-Add and Comp problems, with fixed representation sizes of 50, 100 and 150 nodes. If it is the case that larger solutions are better candidates, and that our neutral rule-sets bias towards neutral growth, then we would expect to see degradation of performance (more evaluations needed) with a size of 50, and improvements (fewer evaluations needed) with a size of 150, over a baseline size of 100. 

\begin{figure}[tp]
\centering
\includegraphics[width=0.44\textwidth]{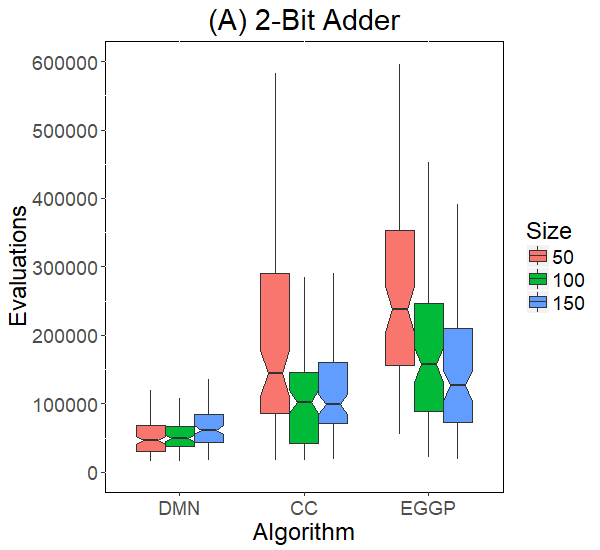}
\includegraphics[width=0.44\textwidth]{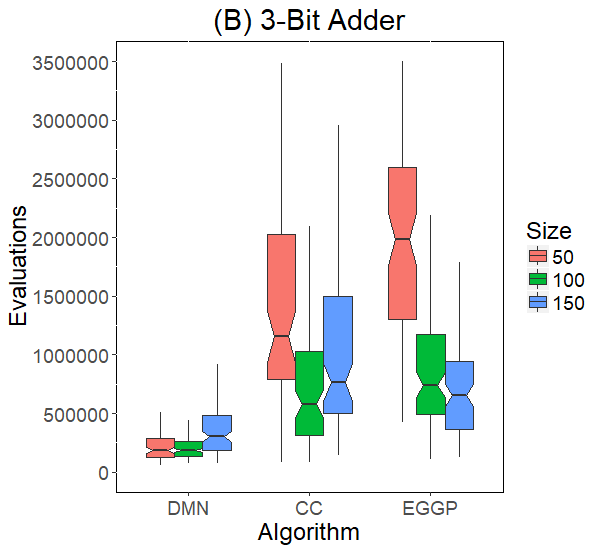}
\includegraphics[width=0.44\textwidth]{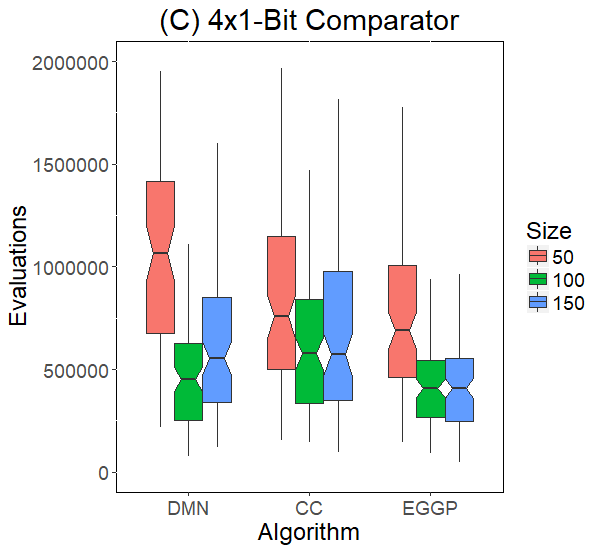}
\caption{Results of running DMN, CC and EGGP on (A) 2-Add, (B) 3-Add and (C) Comp problems. The y axis gives the median evaluations required to solve each problem across 100 runs. The x axis groups setups by algorithm and then lists the observed median evaluations when running that algorithm with 50, 100 or 150 nodes as the fixed representation size. \label{fig:sizeexp}}
\end{figure}

The results of these runs are shown in Fig.~\ref{fig:sizeexp}. For 2-Add and 3-Add with the DMN neutral rule-set, performance actually degrades when increasing the fixed size from $100$ to $150$, while remaining relatively similar when decreasing the size to $50$. For EGGP alone and for the CC neutral rule-set, performance remains relatively similar when increasing the fixed size from $100$ to $150$, but degrades when decreasing the size to $50$. These observations imply that the DMN rule-set is not simply growing solutions to a more beneficial search space, since it performs better when limited to a smaller space. Therefore, on these problems, there is some other property of the DMN rule-set that is benefiting performance.

For the Comp problem, trends remain similar for EGGP alone and the CC neutral rule-set. However, the performance of the DMN rule-set degrades when the fixed size is decreased from $100$ to $50$. This suggests that the Comp problem is in some way different from the other problems. Further, when DMN is run on the Comp problem, the average proportion of active code is nearly 100\%. This may offer an explanation to why the DMN rule-set struggles to outperform standard EGGP on the Comp problem, which has more than twice as many outputs (18) as the next nearest problem (8, DeMux). DMN's bias towards growth paired with the high number of outputs may give some of the problem's many outputs little room to change and configure to a correct solution.

\subsection{DMN and ID in Combination}

We investigate the effect of using DMN and ID,  our two best performing neutral rule-sets, in combination. This combined set, which we refer to as DMID, consists of the following logical equivalence laws:

\begin{center}
DeMorgan$_{F1}$, DeMorgan$_{F2}$, \\
DeMorgan$_{R1}$, DeMorgan$_{R2}$, \\
ID-AND$_F$, ID-AND$_R$, \\
ID-OR$_F$, ID-OR$_R$, \\
ID-NOT$_F$ and ID-NOT$_R$.
\end{center}

We use this set under the same experimental conditions described in Section \ref{sec:Exp} to produce the results given in Table \ref{tab:dmid}. In Table \ref{tab:dmid} we provide $p$ and $A$ values in comparison to the DMN and ID results in Table~\ref{tab:digres} and the EGGP results in Table~\ref{tab:EGGPres}.

\begin{table*}[tp]
\setlength{\tabcolsep}{2pt}
\centering
\begin{tabular}{l rrrccrccrcc}
\toprule
Problem& \multicolumn{2}{c}{DMID} & \, & \multicolumn{2}{c}{vs. DMN} & & \multicolumn{2}{c}{vs. ID} & \, & \multicolumn{2}{c}{vs. EGGP}\\\cline{2-3}\cline{5-6}\cline{8-9}\cline{11-12}
& \multicolumn{1}{c}{ME} & \multicolumn{1}{c}{IQR} & & $p$ & $A$ & & $p$ & $A$ & & $p$ & $A$\\
\midrule
1-Add  & 7,415 & 5,756 & & $10^{-4}$ & 0.64 &  & 0.02 & 0.60  & & $10^{-12}$ & \textbf{0.78}\\
2-Add  & 43,633 & 29,065 & & 0.13 & -- &  & $10^{-8}$ & \textbf{0.73} & & $10^{-23}$ & \textbf{0.91}\\
3-Add  & 162,568 & 112,074 & & 0.02 & 0.60 &  & $10^{-11}$ & \textbf{0.77} & & $10^{-28}$ & \textbf{0.95}\\
\midrule
2-Mul  &  12,020 & 8,761 & & $10^{-3}$ & 0.63 &  & 0.30 & -- & & $10^{-8}$ & \textbf{0.73}\\
3-Mul  &  604,480 & 471,956 & & 0.51 & -- &  & 0.04 & 0.59 & & $10^{-13}$ & \textbf{0.80}\\
\midrule
DeMux  &  20,938 & 11,040 & & $10^{-3}$ & 0.63 &  & $10^{-6}$ & 0.69 & & $10^{-5}$ & 0.68\\
Comp  &  399,140 & 315,459 & & 0.45 & -- &  & $10^{-4}$ & 0.66 & & 0.95 & --\\
\midrule
3-EP  &  3,930 & 3,105 & & $10^{-3}$ & 0.60 &  & $10^{-3}$ & 0.61 & & $10^{-7}$ & 0.71\\
4-EP  & 16,778 & 10,730 & & 0.02 & 0.59 &  & 0.13 & -- & & $10^{-9}$ & \textbf{0.75}\\
5-EP  & 52,868 & 31,445 & & 0.29 & -- &  & $10^{-3}$ & 0.61 & & $10^{-5}$ & 0.66\\
6-EP  & 121,978 & 90,429 & & $10^{-3}$ & 0.61 &  & 0.11 & -- & & $10^{-6}$ & 0.68\\
7-EP  & 326,040 & 224,121 & & 0.95 & -- &  & $10^{-7}$ & 0.70 & & 0.05 & 0.58\\

\bottomrule
\end{tabular}
\caption{Results from Digital Circuit benchmarks for the DMID neutral rule-set. The $p$ value is from the two-tailed Mann-Whitney $U$ test. Where $p<0.05$, the effect size  from the Vargha-Delaney A test is shown; large effect sizes ($A>0.71$) are shown in \textbf{bold}. Statistics are given in comparison to the DMN and ID neutral rule-sets and EGGP.}
\label{tab:dmid}
\end{table*}

The DMID rule-set significantly outperforms DMN on 7 of the 12 problems, and shows no significant difference for the other 5 problems. DMID significantly outperforms ID  on 5 problems (notably the n-Bit Adder problems), shows no significant difference on 3 problems, and is significantly outperformed by ID on 4 problems (notably the 3-Mul, Comp and 7-EP). 
DMID significantly outperforms EGGP without neutral rule-sets on all but 1 problem, with the exception being the Comp problem that DMN also fails to find significant benefits on. These results position DMID and ID on a Pareto front of studied problems, with DMID effectively dominating DMN but neither DMID nor ID universally outperforming each other.

\subsection{\{\texttt{AND}, \texttt{OR}, \texttt{NOT}\}: A Harder Function Set?}

In Table \ref{tab:EGGPres} we show that solving problems with the function set \{\texttt{AND}, \texttt{OR}, \texttt{NOT}\} is significantly more difficult than when using the function set \{\texttt{AND}, \texttt{OR}, \texttt{NAND}, \texttt{NOR}\}. We justify using the former function set over the latter in our experiments as it lends itself to known logical equivalence laws despite costing performance. When we introduce these logical equivalence laws to the evolutionary process with the \{\texttt{AND}, \texttt{OR}, \texttt{NOT}\} function set, this `cost' no longer universally holds. We identify 3-Add, 3-Mul, Comp and 7-EP as the 4 hardest problems, based on the median number of  evaluations required to solve them, Table~\ref{tab:EGGPres}. EGGP with the \{\texttt{AND}, \texttt{OR}, \texttt{NOT}\} function set and augmented with the DMID neutral rule-set significantly ($p < 0.05$) outperforms EGGP with the \{\texttt{AND}, \texttt{OR}, \texttt{NAND}, \texttt{NOR}\} function set on two of the problems. 

These two are the 3-Add ($p = 10^{-10}$, $A = 0.76$) and 3-Mul problems ($p = 10^{-5}$, $A = 0.68$). In contrast, the reverse holds for Comp ($p = 10^{-18}$, $A = 0.85$) and 7-EP ($p = 10^{-14}$, $A = 0.80$). Note that for 3 of these circumstances (excluding 3-Mul), the significant difference occurs with large effect size $(A > 0.71)$.

\begin{figure}[tp]
\centering
\includegraphics[width=0.5\textwidth]{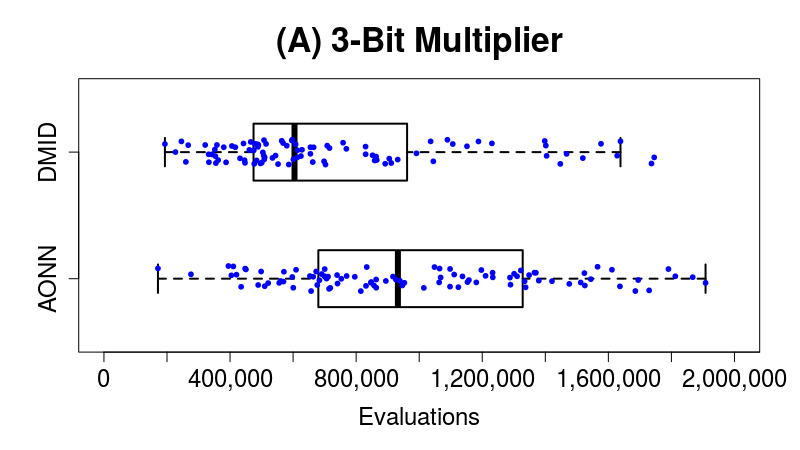}
\includegraphics[width=0.5\textwidth]{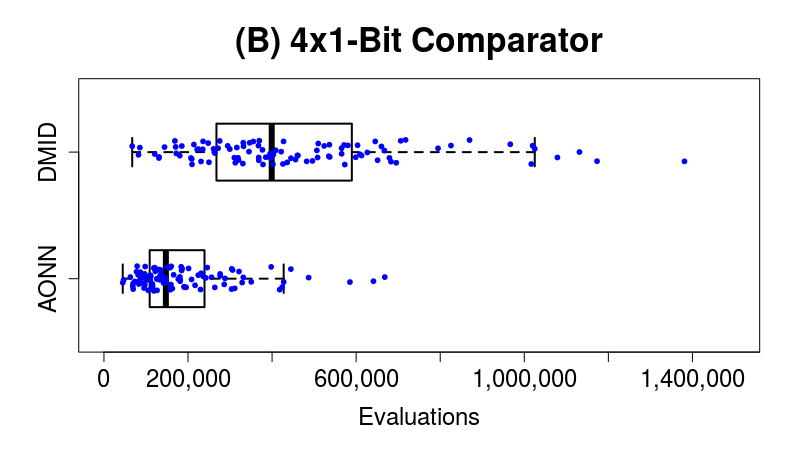}
\caption{Box-plots showing observed evaluations required to solve (A) 3-Bit Multiplier and (B) $4\times1$-Bit Comparator problems using EGGP augmented with the DMID neutral rule-set (DMID) and EGGP with the \{\texttt{AND}, \texttt{OR}, \texttt{NAND}, \texttt{NOR}\} function set (AONN). Vertical jitter is included for visual clarity.\label{fig:boxplot}}
\end{figure}

Fig. \ref{fig:boxplot} shows the number of evaluations across 100 runs for the 3-Mul and Comp problems, for (A) EGGP with the \{\texttt{AND}, \texttt{OR}, \texttt{NOT}\} function set and augmented with the DMID neutral rule-set and (B) EGGP with the \{\texttt{AND}, \texttt{OR}, \texttt{NAND}, \texttt{NOR}\} function set. Here the difference in medians and interquartile ranges for these two evolutionary algorithms can be clearly seen; with EGGP with the DMID neutral rule-set requiring a median evaluations outside of the interquartile range of EGGP with the \{\texttt{AND}, \texttt{OR}, \texttt{NAND}, \texttt{NOR}\} function set for the 3-Mul problem. In stark contrast, the third quartile of evaluations required for the Comp problem lies below the first quartile of EGGP with the DMID neutral rule-set.

This offers an interesting secondary result: there are circumstances and problems where it may be beneficial to choose representations that on their own would yield detrimental results, if that decision then facilitates the inclusion of semantic neutral drift, which may in combination provide enhanced performance over the original representation. 

\section{Conclusions and Future Work \label{sec:Conclusion}}

We have investigated the augmentation of a genetic programming system for learning digital circuits with semantic neutral drift. From our experimental results, we can draw a number of conclusions both for our own specific setting and for the broader evolutionary community.

Firstly, we offer further evidence that there are circumstances where neutral drift aids evolution, building upon existing works that offer evidence in this direction. Additionally, the precise nature of our neutral drift by design offers evidence that neutral drift on the active component of individuals, rather than the intronic components, can aid evolution. For every benchmark problem studied, at least one neutral rule-set was able to yield significant improvements in performance.

Secondly, we have shown that by using graphs as a representation and graph programming as a medium for mutation, it is possible to directly inject domain knowledge into an evolutionary system to improve performance. The application of DeMorgan's logical equivalence laws to graphs with sharing is non-trivial, but becomes immediately accessible in our graph evolution framework. Our ability to design complex domain-specific mutation operators supports the view that that the choice of representation of individuals in an evolutionary algorithm matters. This injection of domain knowledge has been shown to offer benefits beyond simple `neutral growth'.

Thirdly, while the approach we have proposed here offers promising results,  the specific design of neutral drift matters. There are neutral rule-sets that appear to dominate each other, as is found comparing the DMID rule-set to the DMN rule-set. There are also neutral rule-sets which outperform each other on different problems, as is demonstrated comparing the DMID rule-set to the ID rule-set. As we highlighted in comparing DMID to EGGP with what initially appeared to be a preferential function set, there are circumstances where a GP practitioner may want to deliberately degrade the representation in order to access beneficial neutral drift techniques. There are also other circumstances where the cost of incorporating these techniques may outweigh their immediate benefits.

There are a number of immediate extensions to our work that we believe should be investigated. Firstly, the use of the complete function set $\{\texttt{AND}, \texttt{OR}, \texttt{NAND}, \texttt{NOR}, \ttt{NOT}\}$ alongside the DMID semantics preserving mutations and additional mutations for converting between $\texttt{AND}$ and $\texttt{OR}$ gates and their negations via $\ttt{NOT}$ should be investigated. It may be the case that this overall combination yields better results than either of the function sets and semantics preserving mutations we have covered in this work. Additionally, while semantics preserving mutations have generally improved performance with respect to the number of evaluations required to solve problems, it would be worthwhile to measure the clock-time cost of executing these transformations in every generation. Then it would be possible to study the trade-off between gained efficiency and additional overhead. Future work should also investigate the potential use of our proposed approach in CGP and tree-based GP as discussed in Section \ref{sec:snd1}.

While we do not address theoretical aspects of SND here, it may be possible to prove convergence of evolutionary algorithms equipped with SND under certain properties, such as the completeness of the semantics preserving mutations used with respect to equivalence classes. 

There are a number of application domains to investigate for future work: hard search problems where individual solutions may be represented by graphs and where there are known semantics-preserving laws. A primary candidate  is the evolution of Bayesian Network topologies,  a well-studied field \cite{LarranagaKBS13}, as there are known equivalence classes for Bayesian Network topologies \cite{chickering2002learning}. A secondary candidate is learning quantum algorithms using the ZX-calculus, which represents quantum computations as graphs \cite{coecke2011interacting}, and is equipped with graphical equivalence laws that preserve semantics. 

\vspace{5mm}
\renewcommand{\abstractname}{Acknowledgements}
\begin{abstract}
T. Atkinson is supported by a Doctoral Training Grant from the Engineering and Physical Sciences Research Council (EPSRC) in the UK.
\end{abstract}

\bibliographystyle{spmpsci}      

\bibliography{bibliography}

\onecolumn

\appendix

\section{EGGP Programs\label{appendix}}

\subsection{$\texttt{InitCircuit}$\\}

The program $\texttt{InitCircuit}$ in Figure \ref{fig:dac_init} generates EGGP individuals for the digital circuit problems described in this work. This program is defined abstractly, for some function set $F$. The actual form of the first rule-set call is instantiated for a specific function set $F$ where each function $\texttt{f$_\texttt{x}$}$ has a corresponding version of the rule $\texttt{add\char`_node\char`_f$_\texttt{x}$}$ shown in Figure \ref{fig:add_node}. 

This program expects the a problem-specific variant of the graph given in Figure \ref{fig:init_S}, where there are $i$ input nodes and $o$ output nodes and the blue node is labelled with $n$ where $n$ is an integer representing the number of nodes generated individuals should contain. The specific graph in Figure \ref{fig:init_S} will generate circuits with $3$ input nodes, $2$ output nodes and $100$ function nodes.

\begin{figure*}[ht]
\begin{center}
 \begin{tabular}{@{}l@{}}
\small\texttt{Main} := ([\{\texttt{add\_node\_f$_\texttt{x}$ $\mid \texttt{f$_\texttt{x}$} \in F$}\}]; [\texttt{connect\_node}]!; \texttt{unmark\_node})!; [$\texttt{connect\_output}$]!; $\texttt{remove\_counter}$\vspace{5.0mm}\\

\centering
\begin{tabular}{p{7cm}c|cp{7cm}}
\small$\texttt{connect\char`_node(a,b:list; s:string; x:int)}$ 
\medbreak
\centering
\begin{tikzpicture}
\begin{scope}[node distance=3.5cm, every node/.style={rectangle,thick,draw=white,text=black,minimum size=0.6cm}]
    \node (K) {
      \begin{tikzpicture}
      \begin{scope}[node distance=1.1cm, every node/.style={rectangle,thick,draw}]
          \node[label={[xshift=0.45cm, yshift=-0.9cm]:\miniscule1},fill = nodered] (A) {\small$\texttt{a:x}$};
          \node[label={[xshift=0.45cm, yshift=-0.9cm]:\miniscule2},right of =A] (B) {\small$\texttt{s:b}$};
      \end{scope}
      \end{tikzpicture}
 };
 
    \node[right of=K] (R){
      \begin{tikzpicture}
      \begin{scope}[node distance=1.1cm, every node/.style={rectangle,thick,draw}]
          \node[label={[xshift=0.45cm, yshift=-0.9cm]:\miniscule1},fill = nodered] (A) {\small$\texttt{a:x}$};
          \node[label={[xshift=0.45cm, yshift=-0.9cm]:\miniscule2},right of =A] (B) {\small$\texttt{s:b}$};
      \end{scope}
      \begin{scope}[>={Stealth[black]},
                    every node/.style={fill=white,rectangle},
                    every edge/.style={draw,thick}]
                      \path [->, -latex] (A) edge (B);
      \end{scope}
      \end{tikzpicture}
 };
\end{scope}
  \draw[vecArrow] (K) to (R);

  \draw[innerWhite] (K) to (R);
\end{tikzpicture}
\leftskip=0pt\small$\texttt{where s != "OUT" and outdeg(1) < x}$

\bigbreak
\leftskip=0pt\small$\texttt{unmark\char`_node(a:list)}$
\medbreak
\centering
\begin{tikzpicture}
\begin{scope}[node distance=2.5cm, every node/.style={rectangle,thick,draw=white,text=black,minimum size=0.6cm}]
    \node (K) {
      \begin{tikzpicture}
      \begin{scope}[node distance=1.1cm, every node/.style={rectangle,thick,draw}]
          \node[label={[xshift=0.4cm, yshift=-0.9cm]:\miniscule1},fill = nodered] (A) {\small$\texttt{a}$};
      \end{scope}
      \end{tikzpicture}
 };
 
    \node[right of=K] (R){
      \begin{tikzpicture}
      \begin{scope}[node distance=1.1cm, every node/.style={rectangle,thick,draw}]
          \node[label={[xshift=0.4cm, yshift=-0.9cm]:\miniscule1}] (A) {\small$\texttt{a}$};
      \end{scope}
      \end{tikzpicture}
 };
\end{scope}
  \draw[vecArrow] (K) to (R);

  \draw[innerWhite] (K) to (R);
\end{tikzpicture}

& \, & \,&

\small$\texttt{connect\char`_output(s:string; x,y:int)}$ 
\medbreak
\centering
\begin{tikzpicture}
\begin{scope}[node distance=4.5cm, every node/.style={rectangle,thick,draw=white,minimum size=0.6cm,text=black}]
    \node (K) {
      \begin{tikzpicture}
      \begin{scope}[node distance=1.5cm, every node/.style={rectangle,thick,draw,align=center}]
          \node[rectangle,label={[xshift=0.7cm, yshift=-0.9cm]:\miniscule1}] (A) {\small$\texttt{"OUT":x}$};
          \node[rectangle,label={[xshift=0.45cm, yshift=-0.9cm]:\miniscule2},right of =A] (B) {\small$\texttt{s:y}$};
      \end{scope}
      \end{tikzpicture}
 };
 
    \node[right of=K] (R){
      \begin{tikzpicture}
      \begin{scope}[node distance=1.5cm, every node/.style={rectangle,thick,draw,align=center}]
          \node[rectangle,label={[xshift=0.7cm, yshift=-0.9cm]:\miniscule1}] (A) {\small$\texttt{"OUT":x}$};
          \node[rectangle,label={[xshift=0.45cm, yshift=-0.9cm]:\miniscule2},right of =A] (B) {\small$\texttt{s:y}$};
      \end{scope}
      \begin{scope}[>={Stealth[black]},
                    every node/.style={fill=white,rectangle},
                    every edge/.style={draw,thick}]
                      \path [->, -latex] (A) edge (B);
      \end{scope}
      \end{tikzpicture}
 };
\end{scope}
  \draw[vecArrow] (K) to (R);

  \draw[innerWhite] (K) to (R);
\end{tikzpicture}
\leftskip=0pt\small$\texttt{where s != "OUT" and outdeg(1) = 0}$

\bigbreak
\leftskip=0pt\small$\texttt{remove\char`_counter(a:list)}$
\medbreak
\centering
\begin{tikzpicture}
\begin{scope}[node distance=2.5cm, every node/.style={rectangle,thick,draw=white,text=black,minimum size=0.6cm}]
    \node (K) {
      \begin{tikzpicture}
      \begin{scope}[node distance=1.1cm, every node/.style={rectangle,thick,draw}]
          \node[label={[xshift=0.4cm, yshift=-0.9cm]:\miniscule1},fill = nodeblue] (A) {\small$\texttt{a}$};
      \end{scope}
      \end{tikzpicture}
 };
 
    \node[right of=K] (R){
      \begin{tikzpicture}
      \begin{scope}[node distance=1.1cm, every node/.style={rectangle,thick,draw}]
      \end{scope}
      \end{tikzpicture}
 };
\end{scope}
  \draw[vecArrow] (K) to (R);

  \draw[innerWhite] (K) to (R);
\end{tikzpicture}
\end{tabular}
\end{tabular}
\end{center}
\caption{A P-GP\,2 program $\texttt{InitCircuit}$ for generating digital circuits. The program repeatedly probabilistically applies a $\texttt{add$\_$node$\_$f$_\texttt{x}$}$ rule (see Figure \ref{fig:add_node} as long as possible, probabilistically connecting each newly added function node to the existing graph with the $\texttt{connect$\_$node}$ rule until the node's function arity is satisfied. Once the $\texttt{add$\_$node}$ rules are no longer applicable \label{fig:dac_init}, the $\texttt{connect$\_$output}$ rule is applied as long as possible to connect the outputs to the rest of the graph. Finally $\texttt{remove$\_$counter}$ cleans the graph up, removing the blue marked counter node. The generated graph must be acyclic, as edges are only created outgoing from nodes with no incoming edges. }
\end{figure*}
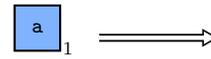

\begin{figure*}[ht]
\begin{center}
\small$\texttt{add\char`_node\char`_f$_\texttt{x}$(n:int)}$ 
\medbreak
\centering
\begin{tikzpicture}
\begin{scope}[node distance=3.5cm, every node/.style={rectangle,minimum size=0.6cm,thick,draw=white,text=black}]
    \node (K) {
      \begin{tikzpicture}
      \begin{scope}[node distance=1.1cm, every node/.style={circle,thick,draw}]
          \node[rectangle,label={[xshift=0.4cm, yshift=-0.9cm]:\miniscule1},fill = nodeblue] (A) {\small$\texttt{n}$};
      \end{scope}
      \end{tikzpicture}
 };
 
    \node[right of=K] (R){
      \begin{tikzpicture}
      \begin{scope}[node distance=2.0cm, every node/.style={circle,thick,draw}]
          \node[rectangle,label={[xshift=0.45cm, yshift=-0.9cm]:\miniscule1},fill = nodeblue] (A) {\small$\texttt{n-1}$};
          \node[rectangle,label={[xshift=1.1cm, yshift=-0.9cm]:\miniscule2},right of =A, fill = nodered] (B) {\small$\texttt{"[f$_\texttt{x}$]":[a$_\texttt{x}$]}$};
      \end{scope}

      \begin{scope}[>={Stealth[black]},
                    every node/.style={fill=white,circle},
                    every edge/.style={draw,thick}]
      \end{scope}
      \end{tikzpicture}
 };
\end{scope}
  \draw[vecArrow] (K) to (R);

  \draw[innerWhite] (K) to (R);
\end{tikzpicture}

\small$\texttt{where n > 0}$
\end{center}
\caption{A P-GP\,2 rule for adding a node of some function $\texttt{f$_\texttt{x}$}$. For the label of node $2$ on the right-hand-side and a specific function $\texttt{f$_\texttt{x}$}$, a unique string representation of $\texttt{f$_\texttt{x}$}$ replaces `$[\texttt{f$_\texttt{x}$}]$' and the arity of $\texttt{f$_\texttt{x}$}$ replaces `$[\texttt{a$_\texttt{x}$}]$'. The blue marked node counter is decreased, and the created function node is marked red so that its edges can be inserted.\label{fig:add_node}}
\end{figure*}
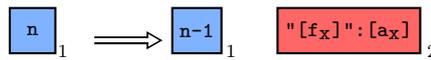

\begin{figure*}[ht]
\centering
\begin{tikzpicture}
\begin{scope}[every node/.style={rectangle,thick,draw,minimum size=0.6cm,}]
    \node[align=center] (1) at (-1, -2) {\miniscule$\texttt{"IN"}:0$};
    \node[align=center] (2) at (1, -2) {\miniscule$\texttt{"IN"}:1$};
    \node[align=center] (3) at (3, -2) {\miniscule$\texttt{"IN"}:2$};
    \node[align=center] (4) at (-1, -3) {\miniscule$\texttt{"OUT"}:0$};
    \node[align=center] (5) at (1, -3) {\miniscule$\texttt{"OUT"}:1$};
    \node[align=center,fill = nodeblue] (6) at (3,-3) {\miniscule$100$};
\end{scope}
\end{tikzpicture}
\caption{The initial graph to be used as input to the program in Figure \ref{fig:dac_init}. Applying the program $\texttt{InitCircuit}$ to this graph will generate acyclic graphs with $3$ inputs, $2$ outputs and $100$ function nodes.\label{fig:init_S}}
\end{figure*}
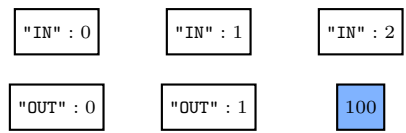

\pagebreak
\subsection{$\texttt{MutateNode}$\\}

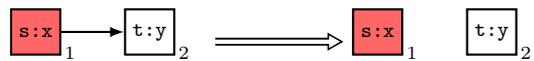
\begin{figure*}[ht]
\begin{center}
 \begin{tabular}{@{}l@{}}
\small\texttt{Main} := [\{\texttt{mutate$\_$node\_f$_\texttt{x}$ $\mid \texttt{f$_\texttt{x}$} \in F$}\}]; \texttt{mark$\_$output}!; [\texttt{add$\_$edge},\texttt{delete$\_$edge}]!; $\texttt{unmark$\_$node}$!\vspace{5.0mm}\\

\centering
\begin{tabular}{p{7cm}c|cp{7cm}}
\small$\texttt{mark\char`_output(a,b,c:list)}$
\medbreak
\centering
\begin{tikzpicture}
\begin{scope}[node distance=3.5cm, every node/.style={rectangle,minimum size=0.6cm,thick,draw=white,text=black}]
    \node (K) {
      \begin{tikzpicture}
      \begin{scope}[node distance=1.1cm, every node/.style={thick,draw}]
          \node[label={[xshift=0.4cm, yshift=-0.9cm]:\miniscule1}] (A) {\small$\texttt{a}$};
          \node[label={[xshift=0.4cm, yshift=-0.9cm]:\miniscule2},right of =A,fill = nodepink] (B) {\small$\texttt{c}$};
      \end{scope}

      \begin{scope}[>={Stealth[black]},
                    every node/.style={fill=white,rectangle,inner sep=0,fill=none},
                    every edge/.style={draw,thick}]
                      \path [->, -latex] (A) edge node[above=0.5pt,inner sep=1pt] {\small$\texttt{b}$} (B);
      \end{scope}
      \end{tikzpicture}
 };
 
    \node[right of=K] (R){
      \begin{tikzpicture}
      \begin{scope}[node distance=1.1cm, every node/.style={thick,draw}]
          \node[label={[xshift=0.4cm, yshift=-0.9cm]:\miniscule1},fill = nodeblue] (A) {\small$\texttt{a}$};
          \node[label={[xshift=0.4cm, yshift=-0.9cm]:\miniscule2},right of =A, fill = nodepink] (B) {\small$\texttt{c}$};
      \end{scope}

      \begin{scope}[>={Stealth[black]},
                    every node/.style={fill=white,rectangle,inner sep=0,fill=none},
                    every edge/.style={draw,thick}]
                      \path [->, -latex] (A) edge node[above=0.5pt,inner sep=1pt] {\small$\texttt{b}$} (B);
      \end{scope}
      \end{tikzpicture}
 };
\end{scope}
  \draw[vecArrow] (K) to (R);

  \draw[innerWhite] (K) to (R);
\end{tikzpicture} 

\bigbreak
\leftskip=0pt\small$\texttt{unmark$\_$node(a:list)}$
\medbreak
\centering
\begin{tikzpicture}
\begin{scope}[node distance=2.5cm, every node/.style={rectangle,minimum size=0.6cm,thick,draw=white,text=black}]
    \node (K) {
      \begin{tikzpicture}
      \begin{scope}[node distance=1.1cm, every node/.style={thick,draw}]
          \node[label={[xshift=0.4cm, yshift=-0.9cm]:\miniscule1},fill = nodepink] (A) {\small$\texttt{a}$};
      \end{scope}
      \end{tikzpicture}
 };
 
    \node[right of=K] (R){
      \begin{tikzpicture}
      \begin{scope}[node distance=1.1cm, every node/.style={thick,draw}]
          \node[label={[xshift=0.4cm, yshift=-0.9cm]:\miniscule1}] (A) {\small$\texttt{a}$};
      \end{scope}
      \end{tikzpicture}
 };
\end{scope}
  \draw[vecArrow] (K) to (R);

  \draw[innerWhite] (K) to (R);
\end{tikzpicture} 

& \, & \,&

\small$\texttt{add\char`_edge(s,t:string; x,y:int)}$ 
\medbreak
\centering
\begin{tikzpicture}
\begin{scope}[node distance=4.5cm, every node/.style={rectangle,thick,draw=white,minimum size=0.6cm,text=black}]
    \node (K) {
      \begin{tikzpicture}
      \begin{scope}[node distance=1.5cm, every node/.style={rectangle,thick,draw,align=center}]
          \node[rectangle,label={[xshift=0.45cm, yshift=-0.9cm]:\miniscule1},fill = nodered] (A) {\small$\texttt{s:x}$};
          \node[rectangle,label={[xshift=0.45cm, yshift=-0.9cm]:\miniscule2},right of =A] (B) {\small$\texttt{t:y}$};
      \end{scope}
      \end{tikzpicture}
 };
 
    \node[right of=K] (R){
      \begin{tikzpicture}
      \begin{scope}[node distance=1.5cm, every node/.style={rectangle,thick,draw,align=center}]
          \node[rectangle,label={[xshift=0.45cm, yshift=-0.9cm]:\miniscule1},fill = nodered] (A) {\small$\texttt{s:x}$};
          \node[rectangle,label={[xshift=0.45cm, yshift=-0.9cm]:\miniscule2},right of =A] (B) {\small$\texttt{t:y}$};
      \end{scope}
      \begin{scope}[>={Stealth[black]},
                    every node/.style={fill=white,rectangle},
                    every edge/.style={draw,thick}]
                      \path [->, -latex] (A) edge (B);
      \end{scope}
      \end{tikzpicture}
 };
\end{scope}
  \draw[vecArrow] (K) to (R);

  \draw[innerWhite] (K) to (R);
\end{tikzpicture}
\leftskip=0pt\small$\texttt{where t != "OUT" and outdeg(1) < x}$

\bigbreak
\leftskip=0pt\small$\texttt{delete\char`_edge(s,t:string; x,y:int)}$ 
\medbreak
\centering
\begin{tikzpicture}
\begin{scope}[node distance=4.5cm, every node/.style={rectangle,thick,draw=white,minimum size=0.6cm,text=black}]
    \node (K) {
      \begin{tikzpicture}
      \begin{scope}[node distance=1.5cm, every node/.style={rectangle,thick,draw,align=center}]
          \node[rectangle,label={[xshift=0.45cm, yshift=-0.9cm]:\miniscule1},fill = nodered] (A) {\small$\texttt{s:x}$};
          \node[rectangle,label={[xshift=0.45cm, yshift=-0.9cm]:\miniscule2},right of =A] (B) {\small$\texttt{t:y}$};
      \end{scope}
      \begin{scope}[>={Stealth[black]},
                    every node/.style={fill=white,rectangle},
                    every edge/.style={draw,thick}]
                      \path [->, -latex] (A) edge (B);
      \end{scope}
      \end{tikzpicture}
 };
 
    \node[right of=K] (R){
      \begin{tikzpicture}
      \begin{scope}[node distance=1.5cm, every node/.style={rectangle,thick,draw,align=center}]
          \node[rectangle,label={[xshift=0.45cm, yshift=-0.9cm]:\miniscule1},fill = nodered] (A) {\small$\texttt{s:x}$};
          \node[rectangle,label={[xshift=0.45cm, yshift=-0.9cm]:\miniscule2},right of =A] (B) {\small$\texttt{t:y}$};
      \end{scope}
      \end{tikzpicture}
 };
\end{scope}
  \draw[vecArrow] (K) to (R);

  \draw[innerWhite] (K) to (R);
\end{tikzpicture}
\leftskip=0pt\small$\texttt{where outdeg(1) > x}$
\end{tabular}
\end{tabular}
\end{center}
\caption{A P-GP\,2 program $\texttt{MutateNode}$ for mutating function nodes of digital circuits. The program probabilistically applies a $\texttt{mutate$\_$node$\_$f$_\texttt{x}$}$ rule (see Figure \ref{fig:mutate_node_fx} to mutate a node's function and mark that node red. In a similar manner to the edge mutation program in Figure \ref{fig:dac_mutate}, all nodes with a directed path to the mutating node are marked blue by $\texttt{mark$\_$output}$ applied as long as possible. Then $\texttt{add$\_$edge}$ and $\texttt{delete$\_$edge}$ can be applied as long as possible to ensure that the node's outgoing edge's respect its new function's arity. Additionally, the fact that all nodes which would introduce a cyclic if tareted are now marked blue ensures that applying $\texttt{add$\_$edge}$ cannot introduce a cycle. Finally $\texttt{unmark$\_$node}$ is used to return the graph to an unmarked state. \label{fig:mutate_node}}
\end{figure*}

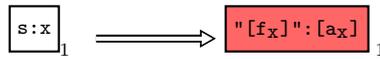
\begin{figure*}[ht]
\begin{center}
\small$\texttt{mutate\char`_node\char`_f$_\texttt{x}$(s:string; x:int)}$ 
\medbreak
\centering
\begin{tikzpicture}
\begin{scope}[node distance=3.5cm, every node/.style={rectangle,minimum size=0.6cm,thick,draw=white,text=black}]
    \node (K) {
      \begin{tikzpicture}
      \begin{scope}[node distance=1.1cm, every node/.style={circle,thick,draw}]
          \node[rectangle,label={[xshift=0.4cm, yshift=-0.9cm]:\miniscule1}] (A) {\small$\texttt{s:x}$};
      \end{scope}
      \end{tikzpicture}
 };
 
    \node[right of=K] (R){
      \begin{tikzpicture}
      \begin{scope}[node distance=2.0cm, every node/.style={circle,thick,draw}]
          \node[rectangle,label={[xshift=1.1cm, yshift=-0.9cm]:\miniscule1},fill = nodered] (A) {\small$\texttt{"[f$_\texttt{x}$]":[a$_\texttt{x}$]}$};
      \end{scope}

      \begin{scope}[>={Stealth[black]},
                    every node/.style={fill=white,circle},
                    every edge/.style={draw,thick}]
      \end{scope}
      \end{tikzpicture}
 };
\end{scope}
  \draw[vecArrow] (K) to (R);

  \draw[innerWhite] (K) to (R);
\end{tikzpicture}

\small$\texttt{where s != "IN" and s != "OUT" and s != "[f$_\texttt{x}$]"}$
\end{center}
\caption{A generic P-GP\,2 rule for mutating a function node to some function $\texttt{f$_\texttt{x}$}$. For the label of node $1$ on the right-hand-side and a specific function $\texttt{f$_\texttt{x}$}$, a unique string representation of $\texttt{f$_\texttt{x}$}$ replaces `$[\texttt{f$_\texttt{x}$}]$' and the arity of $\texttt{f$_\texttt{x}$}$ replaces `$[\texttt{a$_\texttt{x}$}]$'.\label{fig:mutate_node_fx}}
\end{figure*}

The program $\texttt{MutateNode}$ in Figure \ref{fig:mutate_node} mutates EGGP individuals' function nodes for the digital circuit problems described in this work. This program is defined abstractly, for some function set $F$. The actual form of the first rule-set call is instantiated for a specific function set $F$ where each function $\texttt{f$_\texttt{x}$}$ has a corresponding version of the rule $\texttt{mutate\char`_node\char`_f$_\texttt{x}$}$ shown in Figure \ref{fig:mutate_node_fx}. 

\end{document}